\documentclass[journal]{IEEEtran}

\usepackage{epsfig}
\usepackage{graphicx}
\usepackage{amsmath}
\usepackage{amssymb}
\usepackage{caption}
\usepackage{bm}
\usepackage{subfig}
\usepackage{booktabs}
\usepackage{color}
\usepackage{caption} % for color of caption
\usepackage{xcolor} % for color of caption
\usepackage{booktabs} % for color of table
\usepackage{colortbl} % for color of table
\usepackage{verbatim} % for comments
\usepackage{url}

\usepackage{setspace} % for the high of the table
\usepackage{ragged2e} % for alignment of abstract

\begin{document}

\title{High Fidelity Face Manipulation \\ with Extreme Poses and Expressions}
\author{Chaoyou~Fu,
        Yibo~Hu,
        Xiang~Wu,
        Guoli~Wang,
        Qian~Zhang,
        and~Ran~He,~\IEEEmembership{Senior Member,~IEEE}% <-this % stops a space
\IEEEcompsocitemizethanks{\IEEEcompsocthanksitem
C. Fu, Y. Hu, X. Wu, and R. He are with the National Laboratory of Pattern Recognition, CASIA, Center for Research on Intelligent Perception and Computing, CASIA, Center for Excellence in Brain Science and Intelligence Technology, CAS, and the School of Artificial Intelligence, University of Chinese Academy of Sciences, Beijing 100190, China.
E-mail: \{chaoyou.fu, rhe\}@nlpr.ia.ac.cn, \{huyibo871079699,~alfredxiangwu\}@gmail.com. Q. Zhang is with Horizon Robotics, Beijing 100190, China. E-mail: qian01.zhang@horizon.ai. G. Wang is with the Department of Automation, Tsinghua University, Beijing 100084, China. E-mail: wangguoli1990@mail.tsinghua.edu.cn.
(Corresponding author: Ran He.)
}% <-this % stops an unwanted space
}

\markboth{Journal of \LaTeX\ Class Files}%
{Shell \MakeLowercase{\textit{et al.}}: Bare Demo of IEEEtran.cls for IEEE Journals}

\maketitle

\begin{abstract}
Face manipulation has shown remarkable advances with the flourish of Generative Adversarial Networks. However, due to the difficulties of controlling structures and textures, it is challenging to model poses and expressions simultaneously, especially for the extreme manipulation at high-resolution. In this paper, we propose a novel framework that simplifies face manipulation into two correlated stages: a boundary prediction stage and a disentangled face synthesis stage. The first stage models poses and expressions jointly via boundary images. Specifically, a conditional encoder-decoder network is employed to predict the boundary image of the target face in a semi-supervised way. Pose and expression estimators are introduced to improve the prediction performance. In the second stage, the predicted boundary image and the input face image are encoded into the structure and the texture latent space by two encoder networks, respectively. A proxy network and a feature threshold loss are further imposed to disentangle the latent space. Furthermore, due to the lack of high-resolution face manipulation databases to verify the effectiveness of our method, we collect a new high-quality Multi-View Face (MVF-HQ) database. It contains 120,283 images at 6000$\times$4000 resolution from 479 identities with diverse poses, expressions, and illuminations. MVF-HQ is much larger in scale and much higher in resolution than publicly available high-resolution face manipulation databases. We will release MVF-HQ soon to push forward the advance of face manipulation. Qualitative and quantitative experiments on four databases show that our method dramatically improves the synthesis quality.
\end{abstract}

\begin{IEEEkeywords}
Face Manipulation, Extreme Pose and Expression, High-Resolution, MVF-HQ.
\end{IEEEkeywords}

% \fi
\IEEEpeerreviewmaketitle

\section{Introduction} \label{introduction}
\IEEEPARstart{P}{hoto}-realistic face manipulation with poses and expressions is a meaningful task in a wide range of fields, such as movie industry, entertainment, and photography technologies. With the flourish of Generative Adversarial Networks (GANs) \cite{goodfellow2014generative, arjovsky2017wasserstein}, face manipulation has achieved significant advances in recent years~\cite{choi2018stargan, pumarola2018ganimation, shen2018faceid}. However, on the one hand, existing face manipulation methods mainly focus on only one facial variation, \textit{i.e.} only changing poses \cite{huang2017beyond} or expressions \cite{pumarola2018ganimation}. On the other hand, the methods of manipulating large poses and expressions have still been limited to a low-resolution (128$\times$128) \cite{hu2018pose}. Hence, it is challenging to model both poses and expressions \cite{zhang2018joint}, especially for the extreme manipulation (facial poses beyond $\pm 60^o$) at high-resolution.

\begin{figure}[t]
\centering
\includegraphics[width=0.485\textwidth]{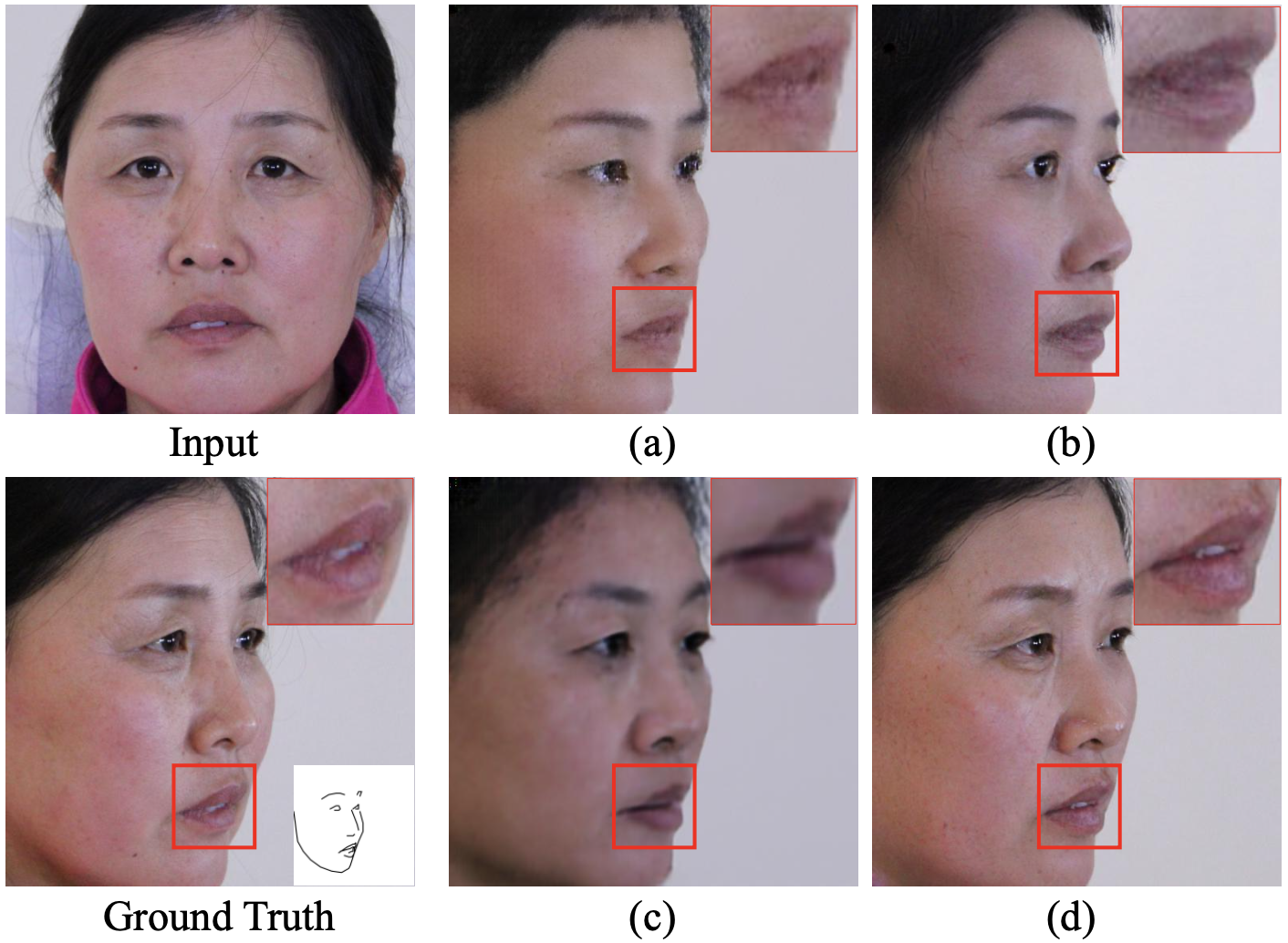}
\caption{Visualization comparisons (512$\times$512 resolution) of different methods. (a) Image-to-image translation \cite{wang2018high}. The local structures, \textit{e.g.} the mouth, are unclear; (b) Concatenating the input face and the boundary of the target face. The local structures are fuzzy and the textures are blurry; (c) Directly utilizing a face recognition network to disentangle structures and textures \cite{bao2018towards}. The local structures are clearer, but the textures are somewhat lost; (d) Our method. Both of the structures and the textures are maintained well.
}
\label{fig-nd-compare}
\end{figure}

For face manipulation, a straightforward way is applying image-to-image translation methods \cite{isola2017image, wang2018high}. However, in the case of extreme manipulation, it is hard for this way to guarantee facial structures and textures. As shown in Fig.~\ref{fig-nd-compare} (a), the facial local structures, such as the eyes, the nose, and the mouth, are unclear. Recent observations \cite{wang2018high, ma2017pose} show that the boundary information is crucial to high fidelity image synthesis.
Meanwhile, several structure-guided methods have been proposed for face manipulation \cite{hu2018pose, jo2019sc}.
For example, CAPG-GAN \cite{hu2018pose} leverages facial landmarks to rotate faces. SC-FEGAN \cite{jo2019sc} realizes facial local editing with sketch guidance. Most structure-guided methods directly concatenate a face image and its structure guidance in the image space. However, such direct concatenation is infeasible for the extreme manipulation at high-resolution, due to more complex structures and textures. As shown in Fig.~\ref{fig-nd-compare} (b), the local structures of the synthesized face are fuzzy and the textures are blurry, such as the mouth.
This phenomenon may be caused by lacking disentanglement, which is important for interpretable image manipulation \cite{shu2018deforming}, between structures and textures. 
Recently, \cite{bao2018towards} proposes a simple disentanglement method. It introduces a face recognition network to extract structure invariant features, and then concatenates the structure invariant features with structure features to synthesize faces. As shown in Fig.~\ref{fig-nd-compare} (c), the disentanglement does make the structure of the synthesized face clearer, but the textures are somewhat lost. 
This may be due to the fact that the features of the face recognition network are too compact, leading to severe texture loss in such a high-resolution case.

Based on the above observations, we propose a novel framework for high-resolution extreme face manipulation, as shown in Fig.~\ref{fig-framework}. Our framework simplifies this challenging task into two correlated stages: a boundary prediction stage and a disentangled face synthesis stage.
The first stage models poses and expressions jointly via boundary images.
Different from most of the previous methods that require manually manipulation of landmarks \cite{ma2017pose,hu2018pose}, we can flexibly generate the desired boundary images merely by controlling the pose and the expression vectors.
Specifically, a conditional encoder-decoder network is employed to predict the boundary image of the target face in a semi-supervised way. Pose and expression estimators are introduced to improve the prediction performance.
The second stage encodes the predicted boundary image and the input face image into the structure and the texture latent space by two encoder networks, respectively. Since it is hard to disentangle structures and textures \cite{shu2018deforming}, we introduce a face recognition network as a proxy to facilitate disentanglement. Instead of leveraging the compacted features of the proxy \cite{bao2018towards} directly, we propose a feature threshold loss to control the compactness between our learned face features and the compacted features. The result of our method in Fig.~\ref{fig-nd-compare} (d) has not only clear structures, but also realistic textures. More synthesis results (1024$\times$1024 resolution) are presented in Fig.~\ref{fig-nd-1024}, from which we observe that our method generates high-quality images even under the largest $90^o$. 
Note that it is rather challenging to manipulate high-resolution faces with such \textbf{extreme facial structure changes}. As far as we know, this has never been achieved except for our work.

\begin{figure}[t]
\centering
\includegraphics[width=0.485\textwidth]{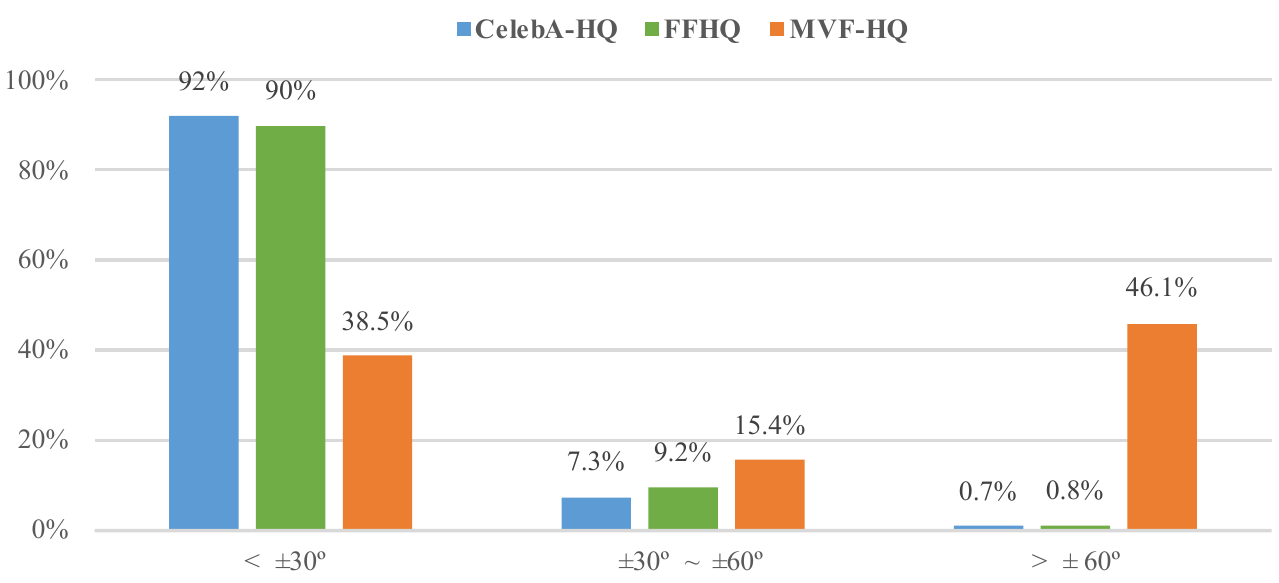}
\caption{Facial pose distribution of high-resolution databases, including CelebA-HQ \cite{karras2017progressive}, FFHQ \cite{karras2018style}, and our MVF-HQ.
}
\label{fig-database}
\end{figure}

In addition, most publicly available face manipulation databases are limited in resolution, \textit{e.g.} the resolution of the original images in the MultiPIE database \cite{gross2010multi} is only 640$\times$480. Although recent high-resolution databases CelebA-HQ \cite{karras2017progressive} and FFHQ \cite{karras2018style} reach a resolution of 1024$\times$1024, the pose variants in these databases are inadequate. 
According to our statistics (estimating poses via the detected facial landmarks), about $92\%$ of the images in CelebA-HQ are nearly frontal (within $\pm 30^o$), and only about $0.7\%$ of the images have large poses (beyond $\pm 60^o$), as shown in Fig.~\ref{fig-database}. For FFHQ, the nearly frontal images and large pose images account for about $90\%$ and $0.8\%$, respectively. Due to the limitation of these databases, it seems impossible for existing GANs-based methods, such as StyleGAN~\cite{karras2018style,karras2019analyzing}, to generate high-quality faces with large poses. Therefore, we collect a new high-quality Multi-View Face (MVF-HQ) that contains 13 views from $-90^o$ to $+90^o$ (the interval is $15^o$), which fills in the blank of the high-resolution multi-view database.
The comparisons between our MVF-HQ database and other publicly available high-resolution face manipulation databases are presented in Fig.~\ref{fig-database} and Table~\ref{table-high-resolution}.
It is obvious that MVF-HQ has the following three advantages:
(1) \textbf{Large-Scale}. MVF-HQ consists of $120,283$ images, far more than other high-resolution databases ($70,000$ at most \cite{karras2018style}).
(2) \textbf{High-Resolution}. The resolution of the original images in MVF-HQ is up to 6000$\times$4000, while other databases only reach 1024$\times$1024 resolution \cite{karras2017progressive, karras2018style}.
(3) \textbf{Abundant Variants}. MVF-HQ contains $13$ views from $-90^o$ to $+90^o$ as well as diverse expressions and illuminations.
It will be released soon to push forward the advance of face manipulation.

In summary, the main contributions are as follows:

\begin{itemize}
  \item High-resolution extreme face manipulation is formulated as a stage-wise learning problem, which contains two correlated stages: a boundary prediction stage and a disentangled face synthesis stage.

  \item The joint modeling of poses and expressions is implemented by the boundary prediction in the first stage, which makes it flexible to obtain desired boundary images. Besides, a proxy network and a feature threshold loss are introduced in the second stage to disentangle structures and textures for realistic synthesis.

  \item We collect a new MVF-HQ database that contains $120,283$ diverse face images at 6000$\times$4000 resolution from $479$ identities. It has much larger scale and much higher resolution than publicly available high-resolution face manipulation databases.

  \item We are the first to achieve the rather challenging high-resolution face manipulation with extreme facial structure changes, even for faces at $90^o$. Extensive experiments on the MultiPIE \cite{gross2010multi}, the RaFD \cite{langner2010presentation}, the CelebA-HQ \cite{karras2017progressive}, and our MVF-HQ databases fully demonstrate the effectiveness of our method.
\end{itemize}

\section{Related Work}

\subsection{Face Manipulation}
Face manipulation has attracted great attention to computer vision and graphics \cite{blanz2003reanimating,wang2009face,yang2011expression,cao2014facewarehouse,kemelmacher2014illumination,chen2019semantic,wang2018video,zakharov2019few,songsri2019face}. Pose rotation \cite{tran2017disentangled, hu2018pose} and expression editing \cite{pumarola2018ganimation, tulyakov2018mocogan} are the two of the main research directions. TP-GAN \cite{huang2017beyond} adopts a two-pathway generative network architecture, achieving photo-realistic face frontalization from a single image. 
DA-GAN \cite{yin2020dual} employs a dual attention mechanism into face frontalization, including a self-attention for the generator and a face-attention for the discriminator.
FaceID-GAN \cite{shen2018faceid} introduces an identity classifier as a competitor to better preserve identity for face synthesis. UV-GAN \cite{deng2018uv} completes the facial UV map to improve the performance of pose-invariant face recognition. By controlling the magnitude of Action Units (AUs), GANimation \cite{pumarola2018ganimation} renders expressions in a continuum. MoCoGAN \cite{tulyakov2018mocogan} separates the hidden features as a content subspace and a motion subspace, which makes it possible to synthesize different expressions. AF-VAE \cite{qian2019make} introduces an additive Gaussian Mixture prior in the structure space, facilitating multi-model face synthesis.

Other face manipulation tasks also have achieved considerable development, such as facial makeup \cite{chang2018pairedcyclegan}, face inpainting \cite{yu2018generative, nazeri2019edgeconnect}, cross-spectral hallucination \cite{lezama2017not,duan2019pose}. PairedCycleGAN \cite{chang2018pairedcyclegan} introduces a new cycle generative network that transfers makeup styles and removes the makeup in an asymmetric manner. By this means, it is able to wear makeup for a target face based on a reference. StarGAN \cite{choi2018stargan,choi2020starganv2} realizes multi-domain face attribute transfer by a single generator.
\cite{yu2018generative} utilizes the surrounding background patches to facilitate image inpainting.
InterFaceGAN \cite{shen2020interpreting} realizes face editing via analyzing the latent semantics encoded by GANs.
\cite{zhang2020unified} first proposes a unified framework to achieve pose-invariant facial expression recognition, face synthesis, and face alignment simultaneously. The elaborate design enables these modules to complement and enhance each other.

Besides the above deep generative model based methods, there are also many approaches based on the 3D morphable model (3DMM) \cite{blanz1999morphable}. As a comparatively mature face analysis technology, 3DMM has been incorporated into various face manipulation tasks. \cite{scherbaum2011computer} suggests the best facial makeup automatically for users. \cite{thies2015real} realizes real-time facial expression transfer via an RGB-D sensor. Face2Face \cite{thies2016face2face} is able to reenact a video sequence according to a source actor. \cite{suwajanakorn2017synthesizing} makes the generated video match a given audio track.

\subsection{Image Synthesis}
As one of the primary means for face manipulation, image synthesis has made great progress in recent years \cite{goodfellow2014generative, kingma2013auto, van2016conditional, li2015generative, dinh2016density}. Image synthesis can be done either in an unconditional manner \cite{karras2017progressive, fu2019dual, miyato2018spectral} or in a conditional manner \cite{isola2017image, park2019semantic}. For unconditional synthesis, images are generated from noises without any conditions. GANs \cite{goodfellow2014generative, radford2015unsupervised, arjovsky2017wasserstein} are the representative unconditional synthesis models that consist of a generator and a discriminator to play a min-max game. The generator synthesizes data from a prior to confuse the discriminator, while the discriminator tries to distinguish the generative data from the real data.
PG-GAN \cite{karras2017progressive} significantly improves the synthesis quality by progressively growing the generator and the discriminator. Variational AutoEncoders (VAEs) \cite{kingma2013auto} are the other representative unconditional synthesis models, which optimize the evidence lower bound objective (ELBO) to learn the data distribution. IntroVAE \cite{huang2018introvae} employs an introspective variational generation model to synthesize high-resolution images without discriminators. VQ-VAE \cite{van2017neural,razavi2019generating} learns discrete representations with an autoregressive prior for high-quality generation.

For conditional image synthesis, the synthesized images need to meet the given conditions. pix2pix \cite{isola2017image} introduces a conditional generative adversarial loss for paired image-to-image translation. pix2pixHD \cite{wang2018high} further improves pix2pix with a coarse-to-fine generator and a multi-scale discriminator, realizing higher fidelity image translation. CycleGAN \cite{CycleGAN2017} proposes a cycle consistent way for unpaired image-to-image translation. BigGAN \cite{brock2018large} first achieves high-resolution (512$\times$512) conditional image synthesis on the ImageNet database \cite{deng2009imagenet}. StyleGAN \cite{karras2018style,karras2019analyzing} employs an alternative generator to learn attributes automatically. Benefitting from the proposed spatially-adaptive normalization, SPADE \cite{park2019semantic} synthesizes photo-realistic landscapes by the semantic layout. 

\subsection{Face manipulation databases}
Publicly available face manipulation databases can be divided into two categories: low-resolution databases and high-resolution databases. Representative low-resolution databases include Celeb-A \cite{liu2015deep}, CAS-PEAL-R1 \cite{gao2007cas}, and MultiPIE \cite{gross2010multi}. 
The resolutions of the original images in these three databases are 505$\times$606, 640$\times$480, and 640$\times$480\footnote{The MultiPIE database contains a small number of frontal images at 3072$\times$2048 resolution, but most of the images are 640$\times$480 resolution.}, respectively. Celeb-A is an in-the-wild face database that contains $202,599$ images with $40$ attribute annotations. It is wildly used in face attribute editing. Both the CAS-PEAL-R1 and the MultiPIE databases, which have $30,863$ and $750,000$+ images respectively, are in the controlled environment. They are mainly adopted for pose invariant face recognition.

Due to the high cost of data acquisition, there are a limited number of high-resolution face manipulation databases. Among them, RaFD \cite{langner2010presentation} is widely used for facial expression analysis. However, both the number of images ($8,040$) and the resolution (681$\times$1024) are limited. 
BioHDD \cite{santos2015biohdd} is mainly leveraged to evaluate biometric identification in the case of heavy degradation. The resolution of the registration images in BioHDD is up to 4368$\times$2912, whereas the number of these images is only $606$\footnote{BioHDD also has many low-resolution images with various attributes. Please refer to \cite{santos2015biohdd} for details.}.
Recently released CelebA-HQ \cite{karras2017progressive} and FFHQ \cite{karras2018style} reach 1024$\times$1024 resolution. The former contains $30$k images that mainly derive from the Celeb-A database. Researchers adopt image processing techniques, such as super-resolution, to convert low-resolution images into higher-resolution ones. The latter consists of $70$k images that are crawled from Flickr. However, according to our statistics, most of the images in CelebA-HQ and FFHQ are nearly frontal, making it hard to edit large poses on these databases.

\section{Method}\label{method}
Given an input face $I^a$, the goal of our method is to synthesize the target face $I^b$, according to a given pose vector $p^b$ and an expression vector $e^b$. In addition, we denote the boundary image of the input face and the target face as $B^a$ and $B^b$, respectively. The face manipulation task is explicitly divided into two stages: a boundary prediction stage and a disentangled face synthesis stage, as shown in Fig.~\ref{fig-framework}. In the rest of this section, we will present them in detail.

\subsection{Boundary Prediction}
\begin{figure}[t]
\centering
\includegraphics[width=0.48\textwidth]{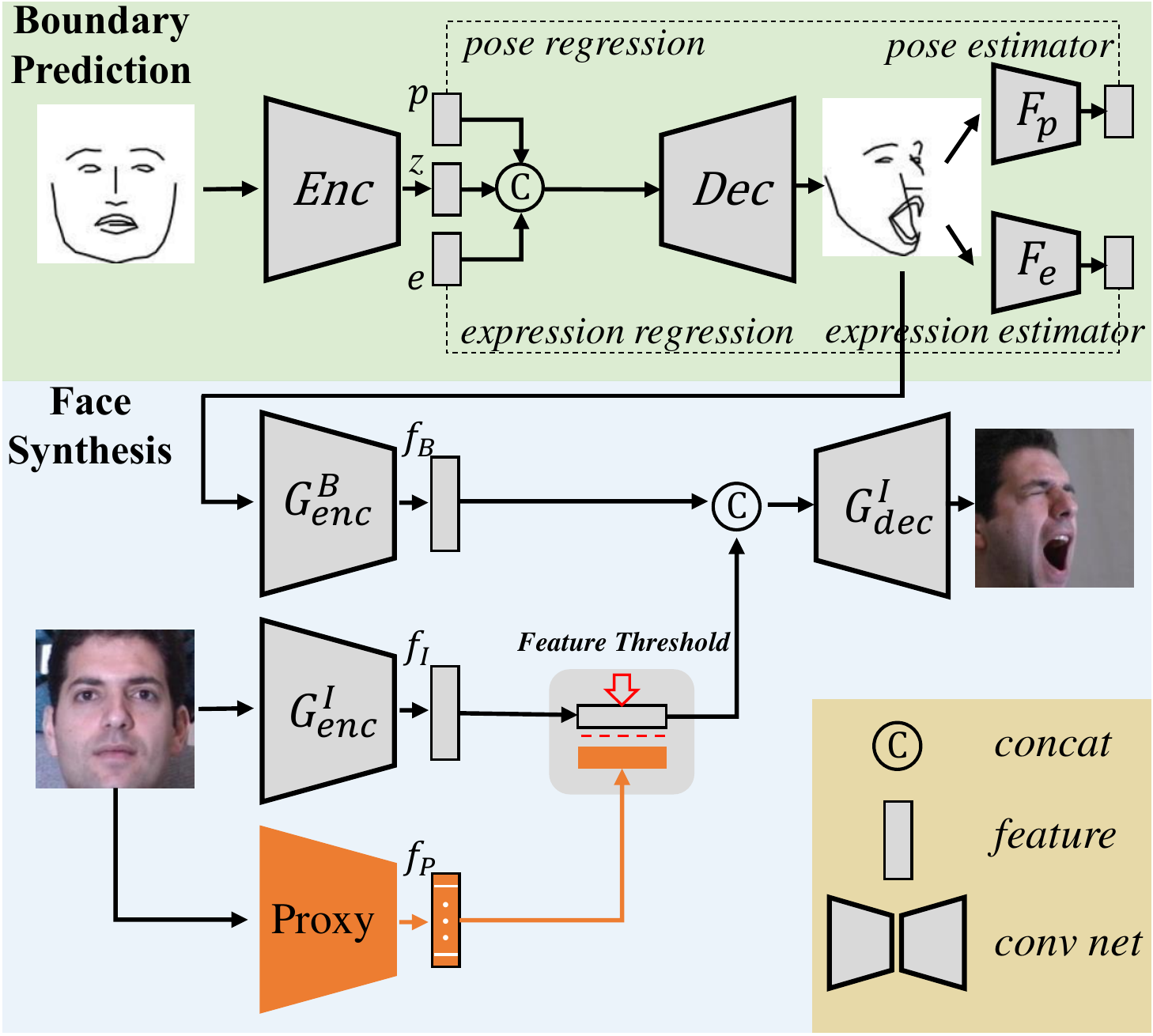}
\caption{The framework of our method, which consists of a boundary prediction stage and a disentangled face synthesis stage. The first stage predicts the boundary image of the target face in a semi-supervised way. Pose and expression estimators are employed to improve the prediction performance. The second stage leverages the predicted boundary image to synthesize the target face. A proxy network and a feature threshold loss are introduced to disentangle structures and textures in the latent space.
}
\label{fig-framework}
\end{figure}

At this stage, we predict the target boundary image according to the given conditional vectors, including a pose vector and an expression vector. As shown in Fig.~\ref{fig-framework}, we utilize an encoder network $Enc$ and a decoder network $Dec$ to realize this conditional boundary prediction. Specifically, through $Enc$, we first map the input boundary image $B^a$ into the latent space $z^a = Enc(B^a)$. Then, the target pose vector $p^b$ and the expression vector $e^b$ are concatenated with the hidden variable $z^a$ to provide conditional information. Lastly, the target boundary image is generated by the decoder network $\hat{B}^b = Dec(z^a, p^b, e^b)$.

The pose and the expression are discrete in the database, \textit{e.g.} the MultiPIE database \cite{gross2010multi} only contains 15 discrete poses and 6 discrete expressions. However, we expect that this stage can generate boundary images with unseen poses and expressions. Hence, we introduce a semi-supervised training manner. For the poses and the expressions in the database, we can utilize the corresponding ground truth to constrain the generated boundary images. For the poses and the expressions that do not exist in the database, we utilize two pre-trained estimators, including a pose estimator $F_p$ and an expression estimator $F_e$, to constrain the generated boundary images by conditional regression.

Two loss functions are involved in this stage, including a pixel-wise loss and a conditional regression loss.

\textbf{Pixel-Wise Loss.} For the poses and the expressions that belong to the database, a pixel-wise $L_1$ loss is utilized to constrain the predicted boundary image $\hat{B}^b = Dec(Enc(B^a), p^b, e^b)$:

\begin{equation}\label{eq:pix-boundary}
    \mathcal{L}_{\text{pix-boud}} = \left | Dec(Enc(B^a), p^b, e^b) - B^b \right |,
\end{equation}
where $B^b$ is the ground truth of the target boundary image.

\textbf{Conditional Regression Loss.} \label{regression}
For the poses and the expressions that do not exist in the database, we first randomly produce $p^r$ and $e^r$ to generate a boundary image $B^r = Dec(z^a, p^r, e^r)$. Then, we utilize a pose estimator $F_p$ and an expression estimator $F_e$, which are pre-trained on the boundary images of the training databases, to estimate the pose $\hat{p}^r = F_p(B^r)$ and the expression $\hat{e}^r = F_e(B^r)$, respectively.
The estimated $\hat{p}^r$ and $\hat{e}^r$ are used to constrain the generated boundary image. The intuition is that the estimated $\hat{p}^r$ and $\hat{e}^r$ of $B^r$ should be equal to the conditional vectors $p^r$ and $e^r$, respectively. Hence, the conditional regression loss, including a pose regression term and an expression regression term, is formulated as:

\begin{equation}\label{eq:regression}
\begin{aligned}
    \mathcal{L}_{\text{reg}} &= || F_p(Dec(z^a, p^r, e^r)) -  p^r||_2^2  \\
    &+ || F_e(Dec(z^a, p^r, e^r)) - e^r ||_2^2.
\end{aligned}
\end{equation}
The parameters of the pre-trained $F_p$ and $F_e$ are fixed during the training procedure.

\subsection{Disentangled Face Synthesis}
This stage leverages the predicted boundary image to perform realistic face synthesis. As shown in Fig.~\ref{fig-framework}, we first utilize two encoders $G_{enc}^{B}$ and $G_{enc}^{I}$ to map the predicted boundary image $\hat{B}^b$ and the input face $I^a$ to $f_{B^b} = G_{enc}^{B}(\hat{B}^b)$ and $f_{I^a} = G_{enc}^{I}(I^a)$, respectively. Then, we disentangle structures and textures in the latent space, by a proxy network $Proxy$ and a feature threshold loss. After the disentanglement, the boundary features $f_{B^b}$ and the image feature $f_{I^a}$ are concatenated to feed into the decoder $G_{dec}^{I}$, synthesizing the final target face $\hat{I^b} = G_{dec}^{I}(f_{B^b}, f_{I^a})$.

The loss functions in this stage are presented as below, including a feature threshold loss, a multi-scale pixel-wise loss, a multi-scale conditional adversarial loss, and an identity preserving loss.

\textbf{Feature Threshold Loss.} \label{feature-threshold-Loss}
The feature threshold loss is designed to assist in disentangling structures and textures in the latent space. Given that directly disentangling is difficult, we employ a face recognition network LightCNN \cite{wu2018light} pre-trained on MS-Celeb-1M \cite{guo2016ms} as a proxy network $Proxy$, whose features $f_{P^a} = Proxy(I^a)$ are thought to be structure invariant \cite{bao2018towards}. In addition, instead of directly utilizing the compact features $f_{P^a}$ that will result in texture loss, as shown in Fig.~\ref{fig-nd-compare} (c), we introduce a feature threshold loss for disentanglement. Specifically, the feature threshold loss controls the feature distance between the learned face features $f_{I^a} = G_{enc}^{I}(I^a)$ and the compact features $f_{P^a} = Proxy(I^a)$ with a threshold margin $m$:

\begin{equation}\label{eq:threshold}
    \mathcal{L}_{\text{thr}} = \left[|| G_{enc}^{I}(I^a) - Proxy(I^a) ||_2^2 - m \right]^{+},
\end{equation}
where $[\cdot]^{+} = max(0,\cdot)$. As the loss $\mathcal{L}_{\text{thr}}$ decreases, the face features $f_{I^a}$ are closer to the compact features $f_{P^a}$, which means structures and textures are better disentangled. Meanwhile, the threshold margin $m$ controls the compact degree of the face features $f_{I^a}$, which is employed to maintain textures. The parameter analysis of $m$ is presented in Section~\ref{parameter-analysis}.

\textbf{Multi-Scale Pixel-Wise Loss.}
We introduce a multi-scale pixel-wise loss to constrain the synthesized face on different scales. Specifically, with the downsampling operation on factors of 2 and 4, we first obtain an image pyramid of 3 scales of the synthesized faces and the ground truth faces, respectively. Then, we calculate the pixel-wise loss on these 3 scales faces:
\begin{equation}\label{eq:pix-2}
    \mathcal{L}_{\text{pix-mul}} = \sum_{s=1,2,3} \left |G_{dec}^{I}(f_{B^b}, f_{I^a})_s - I^b_s \right |,
\end{equation}
where $s$ denotes the scales. The pixel-wise loss at the top of the image pyramid pays more attention to the global information, because it has a larger receptive field. On the contrary, the pixel-wise loss in the bottom of the image pyramid is more concerned with the recovery of details.

\textbf{Multi-Scale Conditional Adversarial Loss.}
To improve the sharpness of the synthesized face image, we also introduce a conditional adversarial loss. The discriminator tries to distinguish the fake image pair $\{\hat{I}^b, B^b \}$ from the real image pair $\{I^b, B^b \}$, while the generator tries to fool the discriminator:
\begin{equation}\label{eq:adv}
\begin{aligned}
    \mathcal{L}_{\text{adv}} &= \mathbb{E}_{I^b \sim P(I^b)} \left [ \log D(I^b, B^b)  \right ] \\
    & + \mathbb{E}_{\hat{I}^b \sim P(\hat{I}^b)} \left [ \log(1 - D(\hat{I}^b, B^b)) \right ].
\end{aligned}
\end{equation}
In order to improve the ability of the discriminator, we adopt the strategy of \cite{wang2018high} that discriminates the synthesized images on three different scales.

\textbf{Identity Preserving Loss.}
In order to further preserve the identity information of the synthesized faces, we adopt an identity preserving loss as \cite{hu2018pose}. Specifically, a pre-trained face recognition network \cite{wu2018light} is introduced as a feature extractor $D_{ip}$. It forces the identity features of the synthesized face $\hat{I}^b$ to be as close to the identity features of the real face $I^b$ as possible. The identity preserving loss is formulated as:
\begin{equation}\label{eq:ip}
    \mathcal{L}_{\text{ip}} = || D_{ip}^{p}(\hat{I}^b) - D_{ip}^{p}(I^b) ||_2^2 + || D_{ip}^{fc}(\hat{I}^b) - D_{ip}^{fc}(I^b) ||_2^2.
\end{equation}
where $D_{ip}^{p}$ and $D_{ip}^{fc}$ denote the output of the last pooling layer and the fully connected layer, respectively.

\subsection{Overall Loss}\label{overal-loss}
The boundary prediction stage and the disentangled face synthesis stage are trained separately. 
That is, the boundary prediction stage is first trained, and then the face synthesis stage is trained based on the predicted boundary.
For the boundary prediction stage, the overall loss is:
\begin{equation}\label{eq:bp}
    \mathcal{L}_{\text{bp}} = \mathcal{L}_{\text{pix-bound}} + \lambda_1 \mathcal{L}_{\text{reg}}.
\end{equation}
For the face synthesis stage, the overall loss is:
\begin{equation}\label{eq:fs}
    \mathcal{L}_{\text{fs}} = \mathcal{L}_{\text{adv}} + \alpha_1 \mathcal{L}_{\text{thr}} + \alpha_2 \mathcal{L}_{\text{pix-mul}} + \alpha_3 \mathcal{L}_{\text{ip}},
\end{equation}
where $\lambda_1$, $\alpha_1$, $\alpha_2$, and $\alpha_3$ are trade-off parameters.

\begin{figure}[t]
    \centering
    \includegraphics[width=0.485\textwidth]{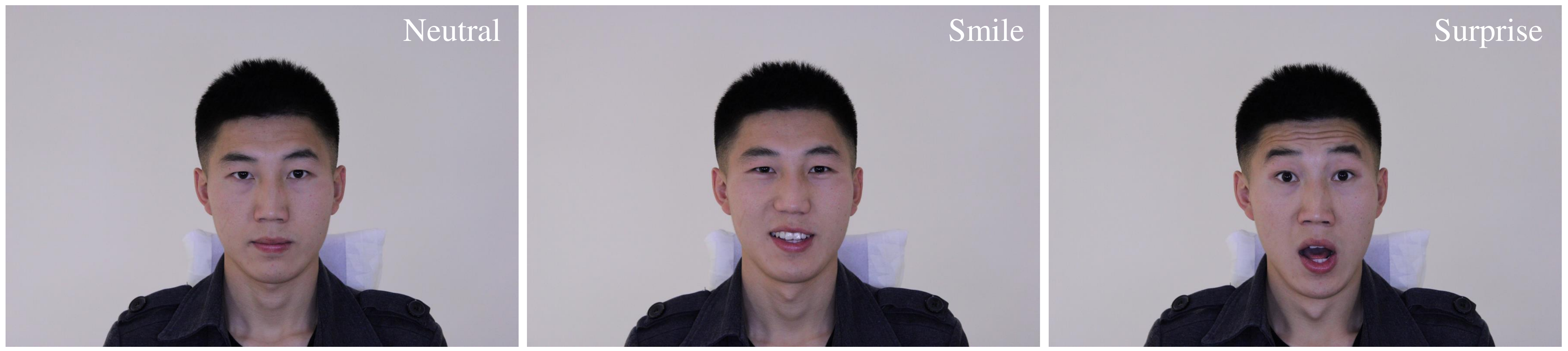}
    \caption{Examples of the expressions.
    }
    \label{fig-expressions}
\end{figure}

\begin{figure}[t]
    \centering
    \includegraphics[width=0.485\textwidth]{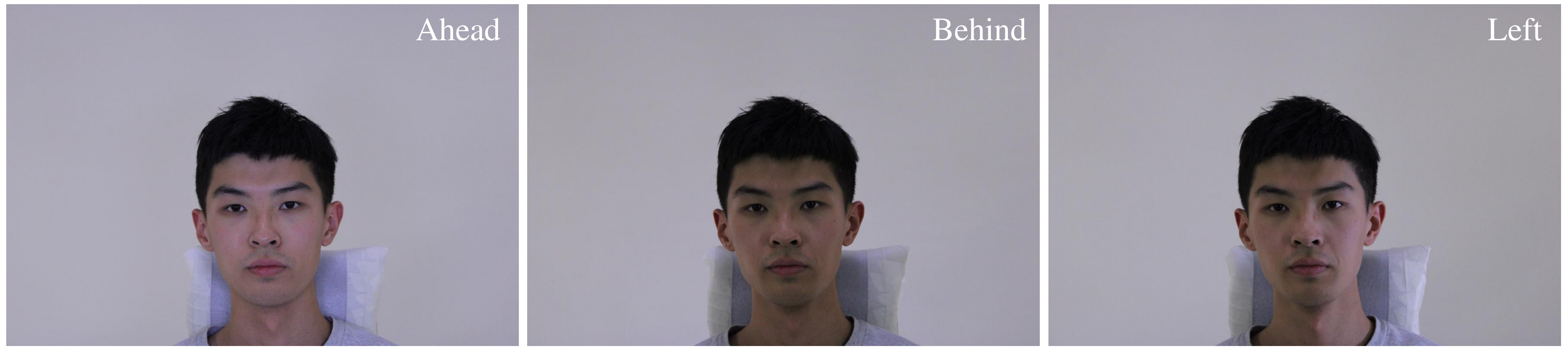}
    \caption{Examples of the illuminations.
    }
    \label{fig-illuminations}
\end{figure}

\begin{figure*}[t]
    \centering
    \includegraphics[width=0.9\textwidth]{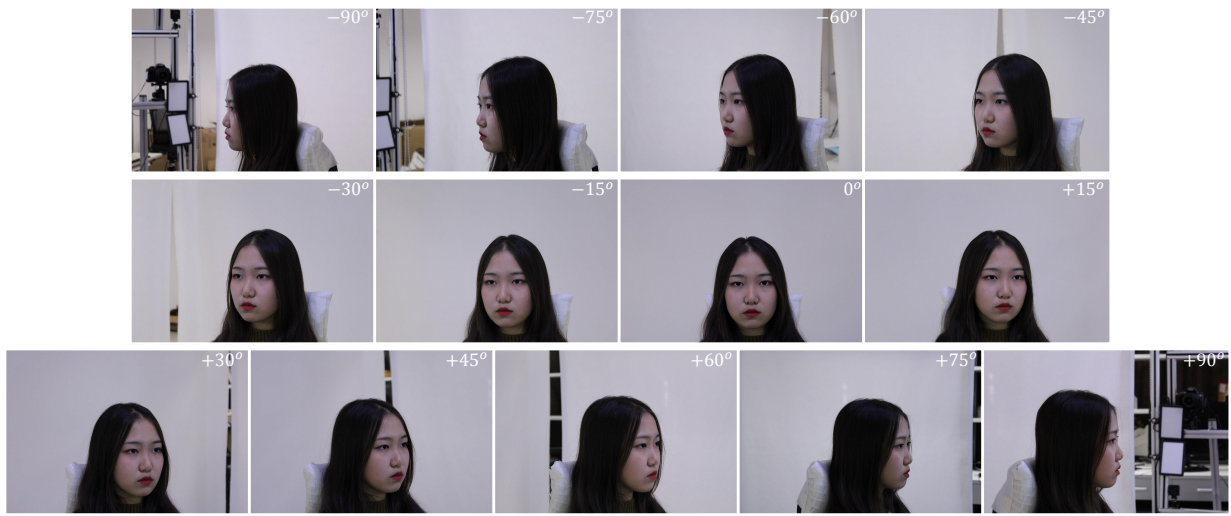}
    \caption{Examples (6000$\times$4000 resolution) of the thirteen views from $-90^o$ to $+90^o$.
    }
    \label{fig-angles}
\end{figure*}

\section{Multi-View Face (MVF-HQ) Database} \label{mvf-hq}
Due to the lack of high-resolution face databases to verify the effectiveness of our method, we collect the MVF-HQ database. This section introduces the details of MVF-HQ in terms of technical setup, data acquisition, data processing, and comparisons.

\subsection{Technical Setup}
$13$ Canon EOS digital SLR cameras (EOS $1300$D/$1500$D with $55$mm prime lens) are equipped to take photos. The resolution of these photos is up to 6000$\times$4000. In order to ensure the accuracy of the poses of the collected photos, we elaborately design and build a horizontal semicircular bracket with a radius of $1.5$m, located at the same height as the head. All cameras are placed on the bracket with $15^o$ interval and point to the center of the semicircular bracket, as shown in Fig.~\ref{fig-setup}.
Meanwhile, all cameras are connected to one computer through USB interfaces.
We design a software that can control all cameras to take photos simultaneously.
The taken photos are automatically stored on the hard drive.

We also use $7$ flashes for illuminations.
These flashes are placed on the above, front, front-above, front-below, behind, left and right, respectively.
By turning on one flash and turning off the others, we can simulate different weak lighting conditions.
Besides, a chair is placed in the center of the semicircular bracket to fix the pose of the participants.
Furthermore, we set a uniform white background for the acquisition environment.

\subsection{Data Acquisition}
\begin{figure}[t]
    \centering
    \includegraphics[width=0.33\textwidth]{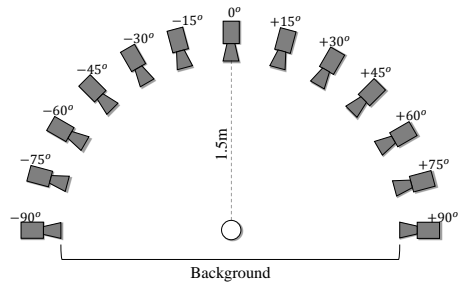}
    \caption{Technical setup.
    }
    \label{fig-setup}
\end{figure}

We invite a total of $500$ participants and all of them have signed data acquisition licenses before taking photos.
Each participant is asked to sit down in the chair, and then fine-tune the height of the chair to ensure the head is at the same height as cameras.
During the process of data acquisition, the participant is asked to look into the direction of the camera on $0^o$ (see Fig.~\ref{fig-setup}) and show three facial expressions, including neutral, smile and surprise, respectively.
Each expression is photographed under all illuminations.
The flashes are switched automatically and quickly to guarantee the consistency of poses and expressions under different illuminations.
The examples of different views, expressions, and illuminations are displayed in Fig.~\ref{fig-angles}, Fig.~\ref{fig-expressions}, and Fig.~\ref{fig-illuminations}, respectively.

\begin{table}[t]
    \centering
    \caption{Comparisons of publicly available low-resolution (the first four databases) and high-resolution (the last five databases) face manipulation databases. `-' means no label.}
    \label{table-high-resolution}
    \begin{spacing}{1.0}
    \resizebox{0.49\textwidth}{!}{
    \begin{tabular}{lcccccccc}
        \toprule[1.0pt]
        Database & Images & Resolution & ID & Poses & Expressions \\
        \midrule[0.9pt]
        PIE \cite{sim2002cmu} & 41,000+ & 640$\times$486 & 68 & 13 & 4 \\
        MultiPIE \cite{gross2010multi} & 750,000+ & 640$\times$480 & 337 & 15 & 6 \\
        CelebA \cite{liu2015deep} & 202,599 & 505$\times$606 & 10,177 & - & - \\
        CAS-PEAL-R1 \cite{gao2007cas} & 30,863 & 640$\times$480 & 1,040 & 21 & 5 \\
        \midrule
        RaFD \cite{langner2010presentation} & 8,040 & ~681$\times$1024 & 73 & 5 & 8 \\
        CelebA-HQ \cite{karras2017progressive} & 30,000 & 1024$\times$1024  & - & - & - \\
        FFHQ \cite{karras2018style} & 70,000 & 1024$\times$1024 & - & - & - \\
        BioHDD \cite{santos2015biohdd} & 606 & 4368$\times$2912 & 101 & 3 & 1 \\
        \midrule
        MVF-HQ & 120,283 & 6000$\times$4000 & 479 & 13 & 3 \\
        \bottomrule[1.0pt]
    \end{tabular}}
    \end{spacing}
\end{table}

\subsection{Data Processing}
After data acquisition, we carefully check each original image to clean the database.
The blurred images and the images with nonstandard poses are removed from the database.
Ultimately, we select $120,283$ images from $479$ identities.
Since it is hard for landmark detection algorithms to accurately detect landmarks under large poses, we manually mark five landmarks for the faces with poses of $\pm60^o$, $\pm75^o$, and $\pm90^o$.
The landmarks of other poses are automatically detected by the algorithm and checked by human.
All facial landmarks will be released along with the MVF-HQ database.

\subsection{Comparisons}
Table~\ref{table-high-resolution} presents the comparisons between our MVF-HQ database and publicly available high-resolution databases.
We observe that MVF-HQ has the following three advantages:
(1) Large-Scale. MVF-HQ consists of $120,283$ images, far more than other databases ($70,000$ at most \cite{karras2018style}).
(2) High-Resolution. The high-performance digital SLR camera enables the obtained images to have a resolution of 6000$\times$4000, while other high-resolution databases only reach 1024$\times$1024 resolution \cite{karras2017progressive, karras2018style}.
(3) Abundant Variants. MVF-HQ contains diverse poses, expressions, and illuminations.

\section{Experiments}
In this section, we evaluate our method on the MultiPIE \cite{gross2010multi}, the RaFD \cite{langner2010presentation}, the CelebA-HQ \cite{karras2017progressive}, and our newly built MVF-HQ databases. The introductions of these databases, such as the number of the images, the resolution, and other attributes, are listed in Table~\ref{table-high-resolution}. The details of these databases and experimental settings are reported in Section \ref{database-and-settings}. In Section \ref{qualitative-experiments} and Section \ref{quantitative-experiments}, extensive qualitative and quantitative results are provided, respectively. In Section \ref{experimental-analysis}, detailed experimental analyses are described.

\begin{figure*}[t]
    \centering
    \includegraphics[width=0.985\textwidth]{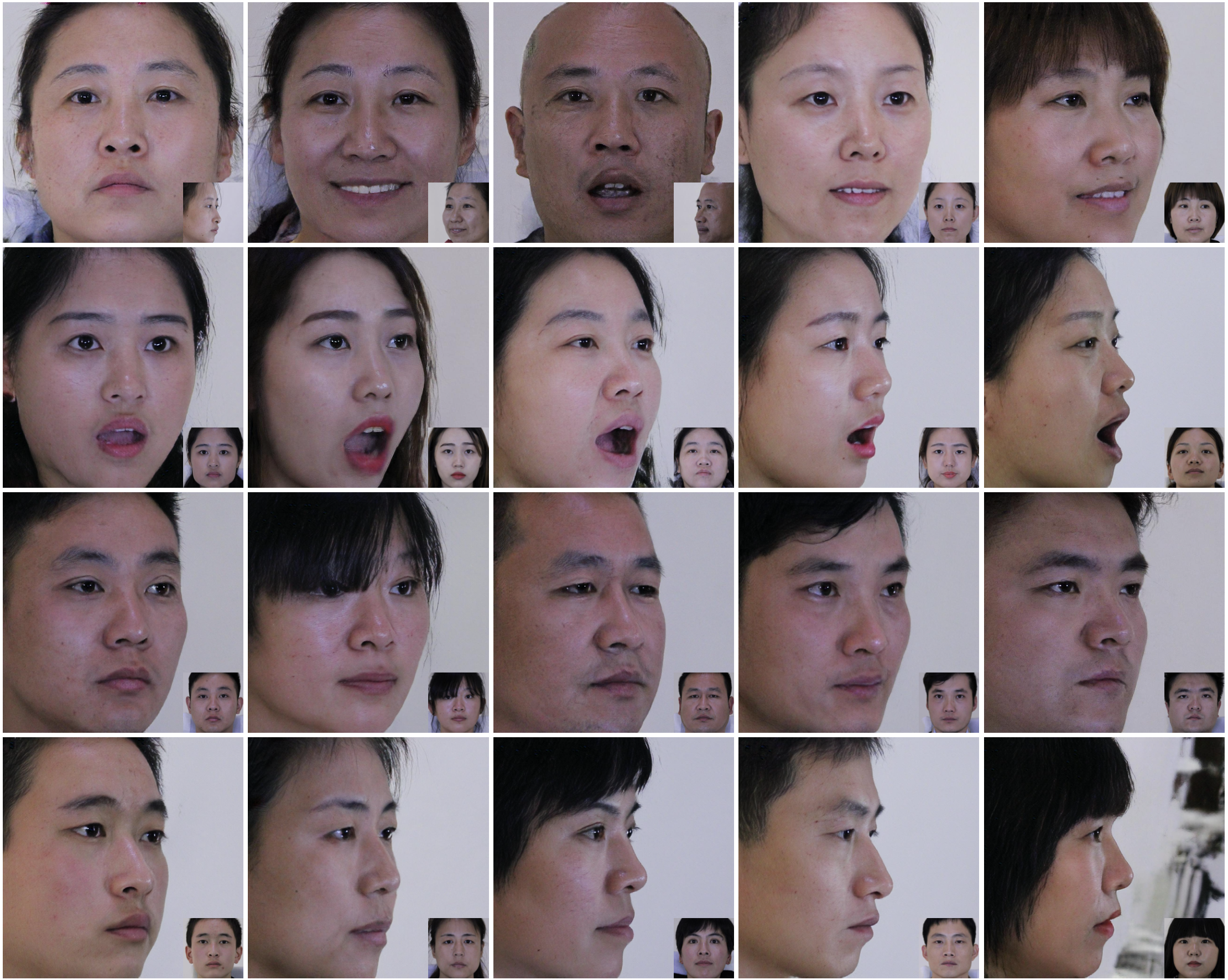}
    \caption{Synthesis results (1024$\times$1024 resolution) on the MVF-HQ database. The lower right corner is the input face.
    }
    \label{fig-nd-1024}
\end{figure*}

\begin{figure*}[t]
    \centering
    \includegraphics[width=0.985\textwidth]{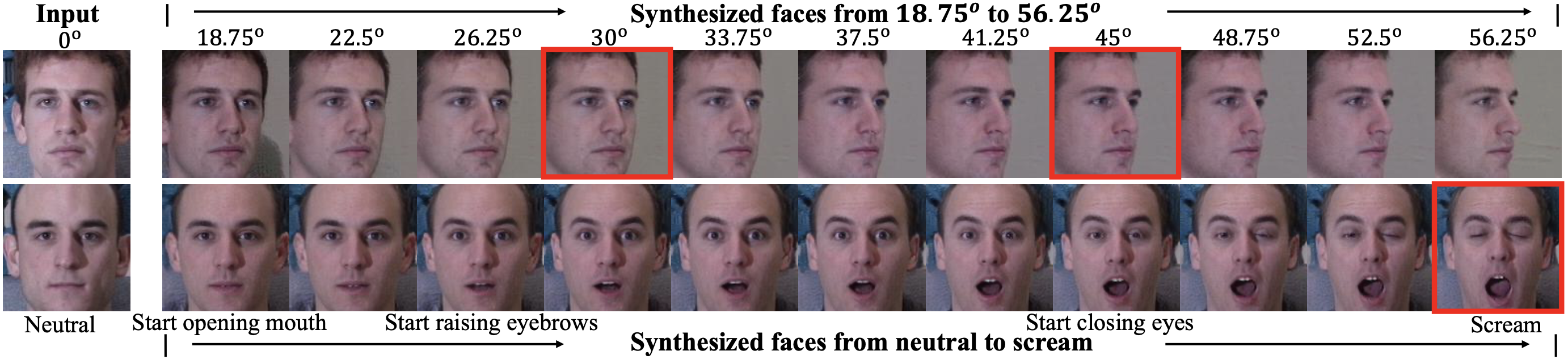}
    \caption{Continuous pose (the first row) and expression (the second row) synthesis on the MultiPIE database. Only the poses and expressions with red boxes are in the database. Please zoom in for details.
    }
    \label{fig-continuous}
\end{figure*}

\begin{figure}[t]
    \centering
    \includegraphics[width=0.485\textwidth]{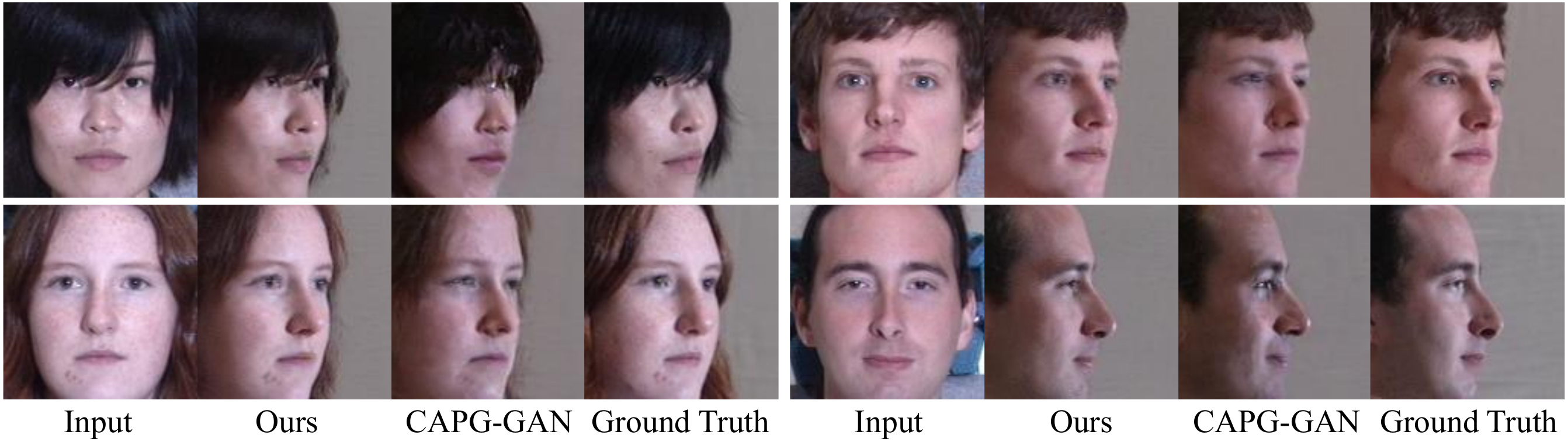}
    \caption{Visualization comparisons between our method and CAPG-GAN \cite{hu2018pose} on the MultiPIE database. For fair comparison, the two methods adopt same boundary images as geometry guidance, the same discriminator, and the same losses except for the feature threshold loss.
    }
    \label{fig-multipie-compare}
\end{figure}

\subsection{Databases and Settings} \label{database-and-settings}
\textbf{MultiPIE} is a low-resolution multi-view database for face recognition and synthesis. In our experiments, we adopt two different settings for quantitative and qualitative experiments, respectively. For the quantitative recognition experiments, following the Setting $2$ protocol of \cite{yim2015rotating, hu2018pose}, we only use the faces with the natural expression. $200$ subjects are used for training and the remaining $137$ subjects are used for testing. For the testing set, the first face of each subject is used as the gallery and the other faces are used as probes. The setting protocol of our qualitative experiments is mainly based on the above Setting $2$ protocol. The difference is that, besides the natural expression, we also use the other $5$ expressions for expression editing. All images are aligned and cropped to 128$\times$128 resolution.

\textbf{RaFD} is a high-resolution face database for expression analysis. It consists of $8$ expressions (see Fig.~\ref{fig-rafd}) and $5$ camera angles ($\pm90^o$, $\pm45^o$, and $0^o$). Furthermore, each identity also contains three different gazed directions (left, frontal, and right). We randomly select $10$ identities as the testing set and use the remaining identities as the training set.
Each image is aligned and cropped to 512$\times$512 resolution.

\textbf{CelebA-HQ} is a high-quality version of the CelebA database \cite{liu2015deep}, containing abundant attributes but with low image quality.
As shown in Fig.~\ref{fig-database}, most of the faces in CelebA-HQ are nearly frontal, making it hard to edit large pose faces. In order to enrich the poses of CelebA-HQ, we utilize a $3$D model \cite{zhu2016face} to create profiles. 
In addition, given that the created profiles have many artifacts, as shown in Fig.~\ref{fig-celeba-hq}, we only conduct face frontalization experiments, \textit{i.e.} rotating profiles to the frontal view. We randomly choose $3,000$ images as the testing set and use the remaining images as the training set. Each image is resized to 512$\times$512 resolution.
Furthermore, considering the abundant pose variants in the CelebA database, we also verify the effectiveness of our method on it. $2000$ images are selected as the testing set. Each image is aligned and cropped to 128$\times$128 resolution.

\textbf{MVF-HQ} is the newly built high-resolution multi-view face database, the details of which are reported in Section \ref{mvf-hq}. In experiments, we randomly select $336$ identities as the training set and the remaining $143$ identities as the testing set. There are no identity overlaps between training and testing. In addition, due to the limited GPU memory, we only conduct experiments at 512$\times$512 and 1024$\times$1024 resolutions.

\textbf{Experimental Settings.} \label{experimental-details}
For the boundary image, we use an open-source toolkit \cite{bulat2017far} to automatically detect $68$ landmarks, and manually check and revise the landmarks with extreme poses ($\pm 60^o$, $\pm 75^o$, and $\pm 90^o$) to ensure the accuracy. The adjacent landmarks are connected to get the boundary image that is in RGB format, as shown in Fig.~\ref{fig-nd-compare}. The boundary image mainly consists of five facial components, including eyebrows, eyes, nose, mouth, and jaw. These components can clearly present the pose, the expression, and the shape of faces. The pose vectors are directly calculated according to the detected facial landmarks. The expression vectors are denoted by Action Units (AUs) \cite{friesen1978facial}, which are collected by an open-source toolkit \cite{baltrusaitis2018openface}. The pose estimator $F_p$ and the expression estimator $F_e$ in Section~\ref{regression} are pre-trained on the above four databases as well as the large-scale in-the-wild database CelebA \cite{liu2015deep}.
The dimensionalities of the latent vector $z$, the pose vector $p$, and the expression vector $e$ in Eq.~(\ref{eq:regression}) are $128$, $3$, and $17$, respectively.
Our discriminator is the same as \cite{wang2018high}.
The parameters $\lambda_1$, $\alpha_1$, $\alpha_2$, and $\alpha_3$ in Section~\ref{overal-loss} are set to $0.1$, $0.01$, $50$, and $0.02$, respectively.
The parameter $m$ in Eq.~(\ref{eq:threshold}) is set to $7$. Adam \cite{kingma2014adam} ($\beta_1$ = 0.9, $\beta_2$ = 0.999) is adopted as the optimizer with a fixed learning rate $0.0002$. 
The high-resolution experiments on MVF-HQ are conducted on $8$ NVIDIA Titan RTX GPUs with $24$GB memory. Training takes about $12$ days for 1024$\times$1024 resolution and about $7$ days for 512$\times$512 resolution.

\subsection{Qualitative Experiments} \label{qualitative-experiments}
\begin{figure}[t]
    \centering
    \includegraphics[width=0.485\textwidth]{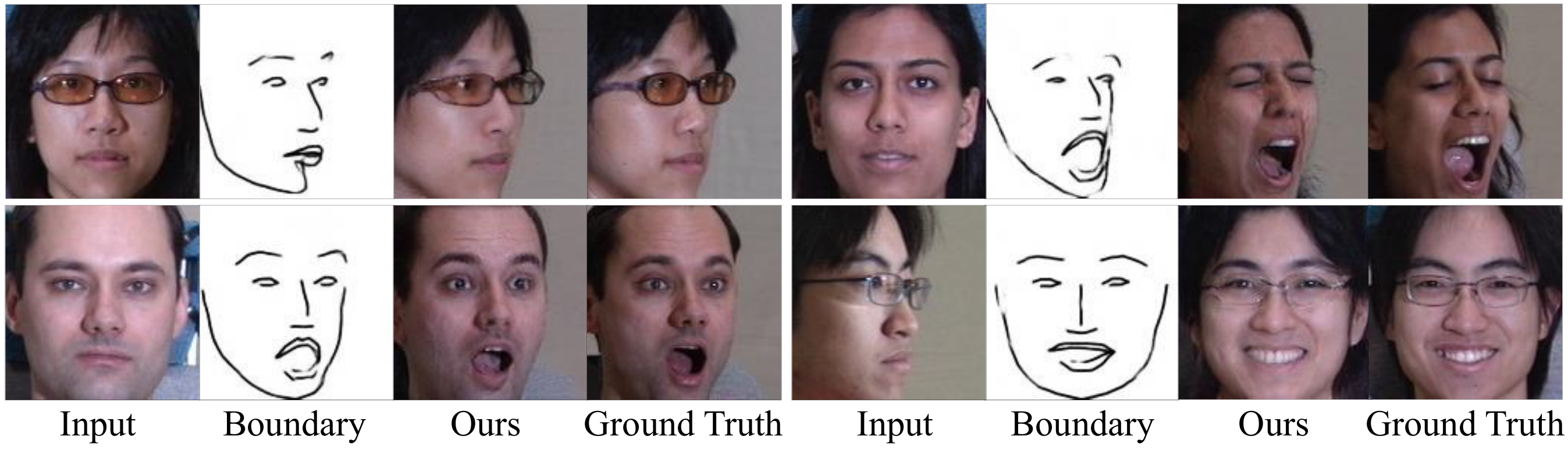}
    \caption{Synthesis results on the MultiPIE database. The boundary is generated in our boundary prediction stage.
    }
    \label{fig-multipie}
\end{figure}

\textbf{Experimental Results on the MultiPIE database.}
According to the given conditional vectors, our method can render an input face to the corresponding poses and expressions. Another state-of-the-art work to tackle the similar task is CAPG-GAN \cite{hu2018pose}, which rotates a face to the target pose by controlling $5$ facial landmarks. 
For fair comparisons, the $5$ facial landmarks are replaced with the same boundary images as our method. Moreover, the discriminator and the losses (except for the feature threshold loss) are also unified.
The comparison results between our method and CAPG-GAN on the MultiPIE database are shown in Fig.~\ref{fig-multipie-compare}. CAPG-GAN concatenates the input faces and the target facial landmarks in the image space, and then feeds them into the generator. As mentioned in Section \ref{introduction}, such a concatenation manner cannot hold the textures well, which is also embodied in Fig.~\ref{fig-multipie-compare}. For example, the details of eyes are somewhat blurry. On the contrary, the synthesized images of our method are closer to the ground truth in terms of both structures and textures. We owe the superiority of our method over CAPG-GAN to the effective disentanglement in the latent space.
In addition, for CAPG-GAN, the target landmarks are manually given if they want to manipulate the input faces. Such a manual manner is difficult and time-consuming. Differently, our method can flexibly generate desired boundary images with the specified pose and expression vectors, as shown in Fig.~\ref{fig-multipie}.
Another advantage of our method over CAPG-GAN is that, besides rendering poses, we can also edit facial expressions (see Fig.~\ref{fig-multipie}).
We observe that the structures of the synthesized faces are consistent with the boundary images. At the same time, the textures of the synthesized images are preserved well, even under the extreme case.

\begin{figure*}[t]
    \centering
    \includegraphics[width=0.985\textwidth]{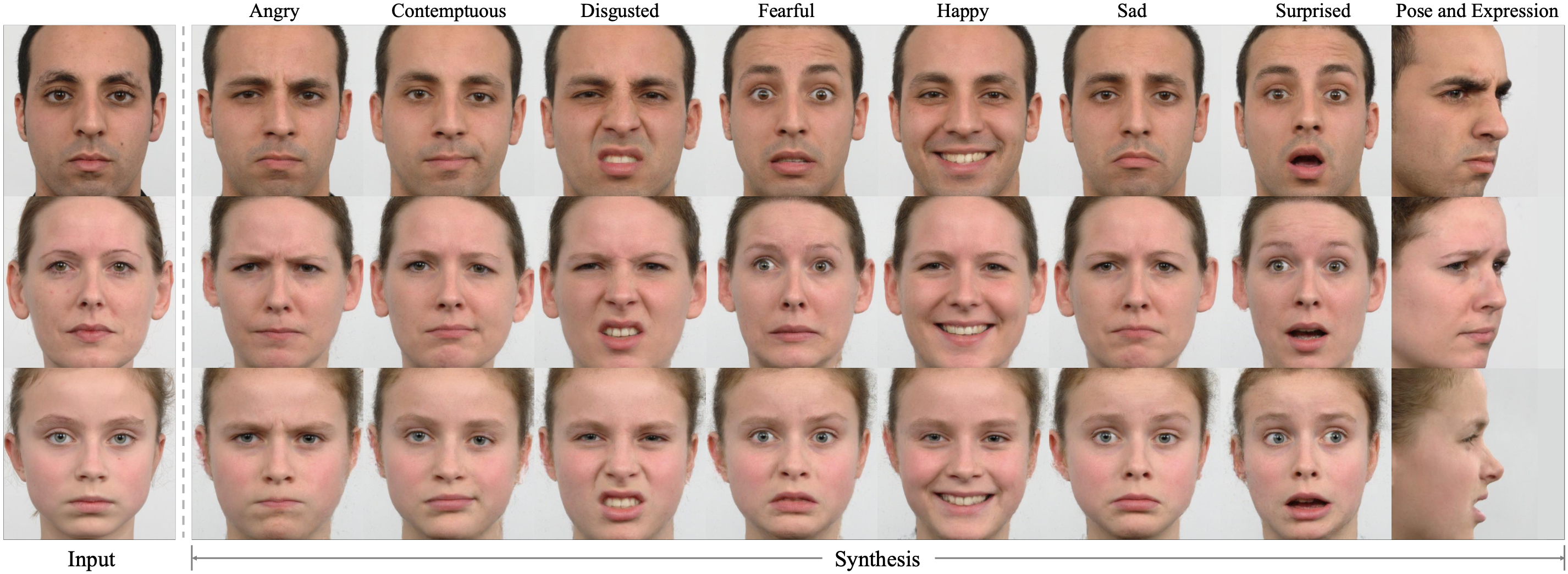}
    \caption{Facial expression and pose synthesis (512$\times$512 resolution) on the RaFD database. The first column is the input, and the remaining columns are synthesis results with different expressions and poses.
    }
    \label{fig-rafd}
\end{figure*}

\begin{figure}[t]
    \centering
    \includegraphics[width=0.44\textwidth]{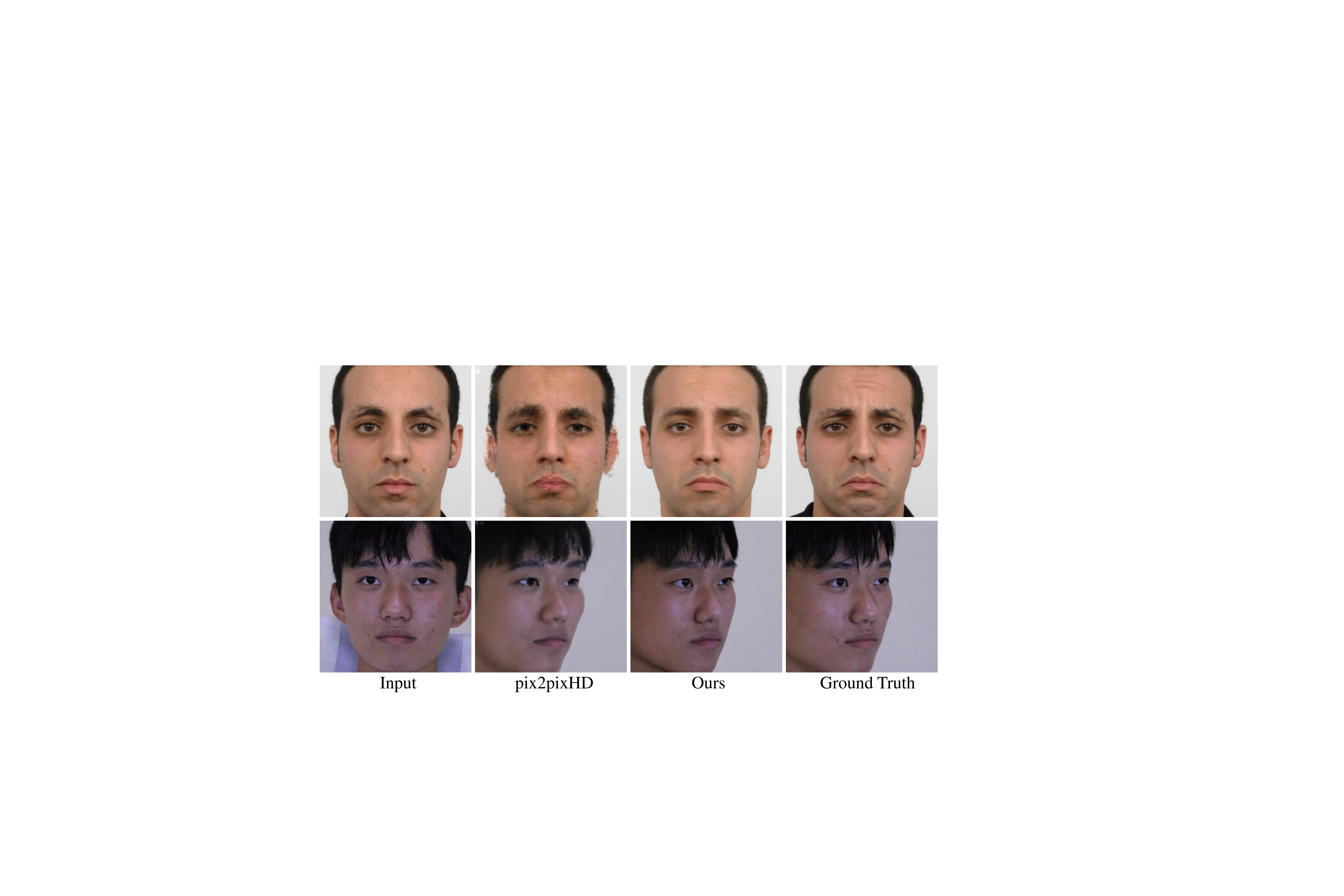}
    \caption{Visualization comparisons (512$\times$512 resolution) with pix2pixHD \cite{wang2018high} on the RaFD database (the first row) and the MVF-HQ database (the second row).
    }
    \label{fig-nd-rafd-compare}
\end{figure}

Fig.~\ref{fig-continuous} further presents the continuous pose and expression synthesis.
All synthesized poses and expressions except for the three images with red boxes are unseen in the training stage. 
The first row displays the results of the continuous rotation from $18.75^o$ to $56.25^o$. The angle interval of the synthesized images is $3.75^o$, while the angle interval in the MultiPIE database is $15^o$. In addition, the second row shows the continuous expression variation from neutral to screaming. The synthesized expressions are quite vivid. The second person with neutral expression begins with opening the mouth gradually. Then, the eyebrows are raising and the eyes begin to close. Ultimately, he makes a scream expression.
The results of the continuous variation demonstrate the generalization ability of our method.

\textbf{Experimental Results on the RaFD database.}
We compare our method with pix2pixHD \cite{wang2018high}, which is a state-of-the-art high-resolution conditional image-to-image translation method. The first row of Fig.~\ref{fig-nd-rafd-compare} plots the expression manipulation results. We observe that, although this task only needs to make slight facial changes, pix2pixHD fails to maintain structures and textures well. The reason may be that pix2pixHD does not disentangle structures and textures. Contrastively, our method gets much better synthesis results. Moreover, in Fig.~\ref{fig-rafd}, we also show the synthesis results of angry, contemptuous, disgusted, fearful, happy, sad, and surprised expressions, respectively. Each synthesized expression is vivid and matches the target expression label. Besides, in the last column of Fig.~\ref{fig-rafd}, we also present the high-quality synthesis results of the joint pose and expression variation.
The high quality of the synthesized images indicates the good performance of our method.

\begin{figure}[t]
    \centering
    \includegraphics[width=0.485\textwidth]{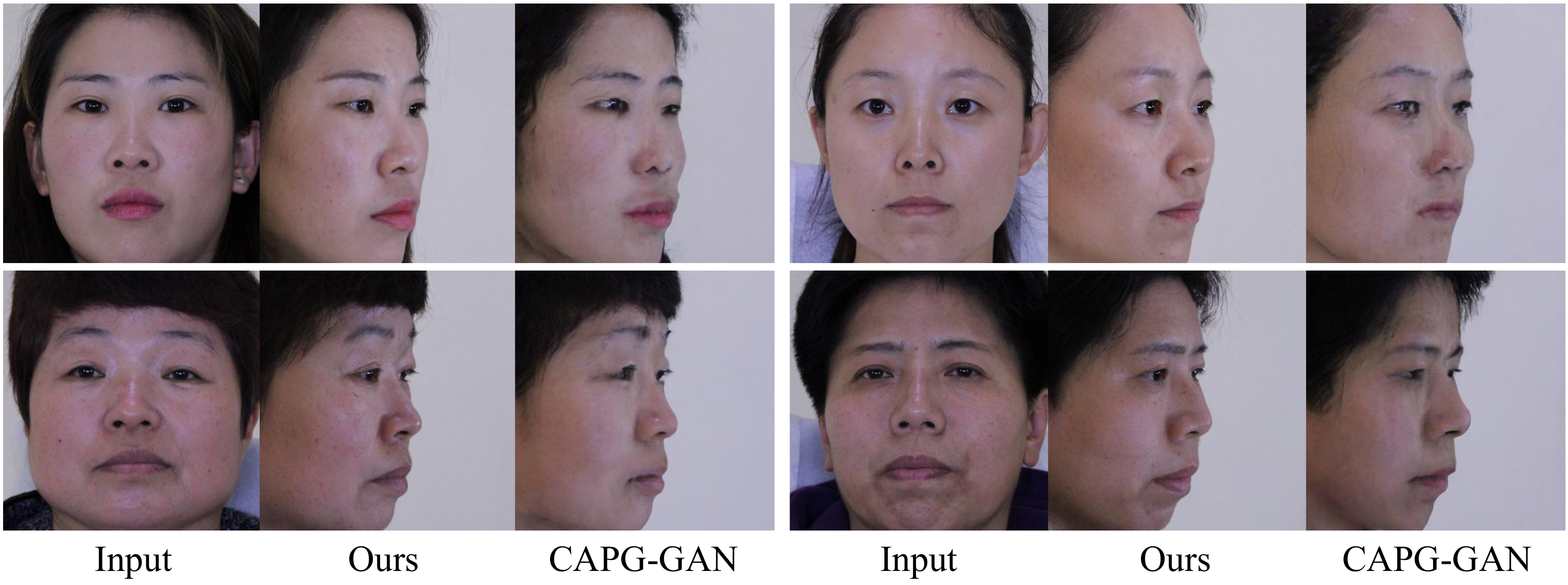}
    \caption{Visualization comparisons (512$\times$512 resolution) between our method and CAPG-GAN on the MVF-HQ database. Please zoom in for details.
    }
    \label{fig-compare-capg-mvf}
\end{figure}

\begin{figure*}[t]
    \centering
    \includegraphics[width=0.985\textwidth]{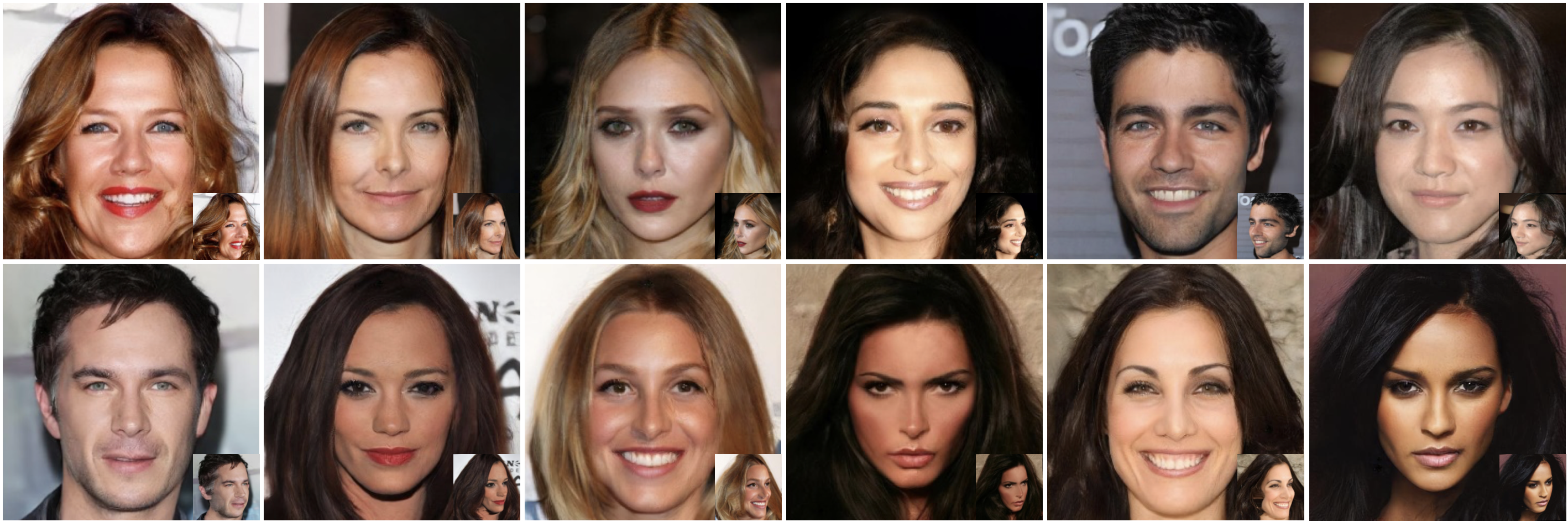}
    \caption{Synthesis results (512$\times$512 resolution) on the CelebA-HQ database. The lower right corner shows the created profile.
    }
    \label{fig-celeba-hq}
\end{figure*}

\begin{figure}[t]
    \centering
    \includegraphics[width=0.485\textwidth]{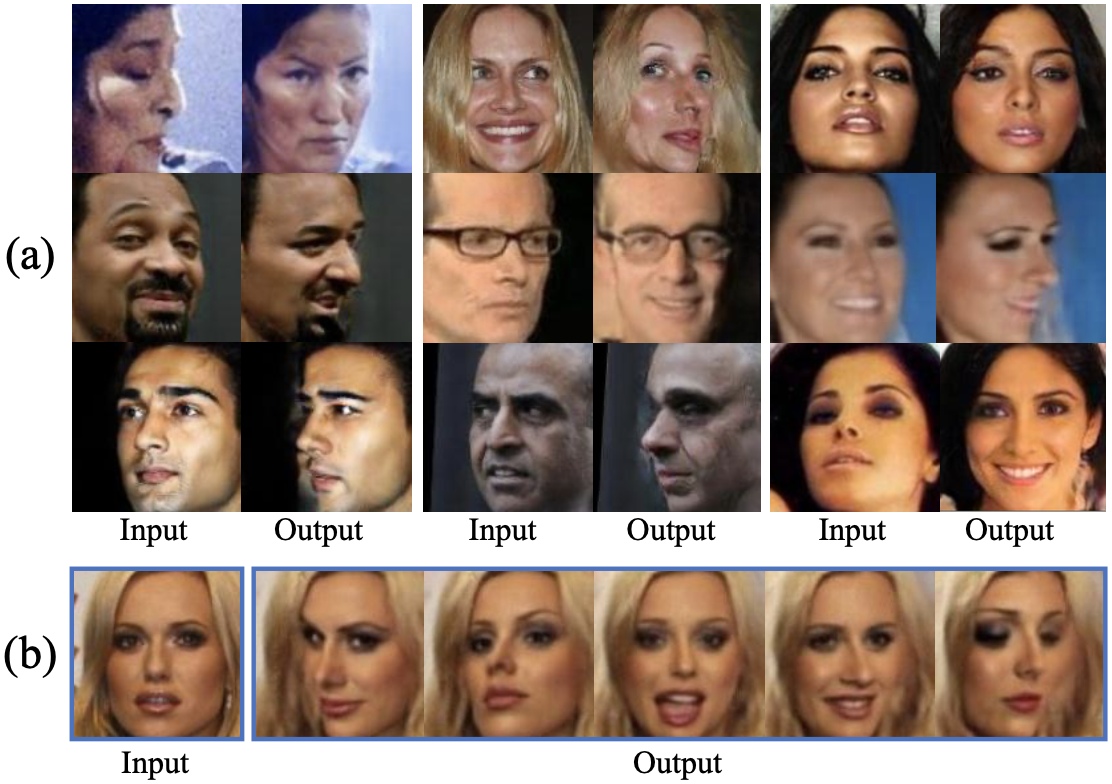}
    \caption{Synthesis results on the CelebA database. (a) Diverse face manipulation results. (b) Multiple manipulations of one input image.
    }
    \label{fig-celeba}
\end{figure}

\textbf{Experimental Results on the MVF-HQ database.}
The second row of Fig.~\ref{fig-nd-rafd-compare} plots the visualization comparisons between our method and pix2pixHD. For the synthesized image of pix2pixHD, the local structures, such as the outline of eyes, are unclear, and the facial textures are blurry.
Conversely, our synthesized image has clear structures and realistic textures.
Furthermore, we discover that the result of our method on the MVF-HQ database is much better than that on the RaFD database. This is because that the MVF-HQ database contains more data (about 120k images) than the RaFD database (about 8k images). It is rather challenging to synthesize high-resolution results with limited training images.

Fig.~\ref{fig-compare-capg-mvf} shows the comparison results between our method and CAPG-GAN on the MVF-HQ database. Same as the experiments on the MultiPIE database, we also unify the the geometry guidance, the discriminator, and the losses. We can see that our method displays significant advantages than CAPG-GAN in such high-resolution cases. Compared with our results, the synthesis results of CAPG-GAN have neither clear local structures (\textit{e.g.} the structure of the eyes) nor realistic textures, especially at large poses. These results demonstrate that it is really infeasible to directly apply the existing low-resolution method to the high-resolution extreme face manipulation.

In Fig.~\ref{fig-nd-1024}, we present more synthesis results (1024$\times$1024 resolution) of different poses and expressions on the MVF-HQ database. We observe that our method obtains satisfactory results, such as recovering the unseen ears of the third face in the first row. More synthesized details, such as the double eyelids of the first face image in the third row, demonstrate the superiority of our method.

\textbf{Experimental Results on the CelebA-HQ database.}
All the above face manipulation databases are in the controlled environment. In order to explore the expansibility of our method under the in-the-wild situation, we conduct experiments on the CelebA-HQ database. As stated in Section \ref{database-and-settings}, most of the images in CelebA-HQ are nearly frontal. The profiles are created by a $3$D model and have too many artifacts. In this case, we only conduct face frontalization experiments. Due to the effects of uncontrolled variants, such as diverse illuminations and backgrounds, it is challenging to perform high-resolution face frontalization under the in-the-wild setting. Fig.~\ref{fig-celeba-hq} shows the results of the synthesized frontal faces. Although the created profiles have massive artifacts and lose many facial textures, our method successfully eliminates the artifacts and completes the lost textures.

We further perform experiments on the original version of the CelebA-HQ database, \textit{i.e.} the CelebA database \cite{liu2015deep} that has abundant poses but with lower image quality. Fig.~\ref{fig-celeba} (a) plots the manipulation results with diverse poses, from which we observe that our method has the ability to rotate extreme poses, such as the first set of images. 
Nevertheless, since the training set of the CelebA database has massive low-quality images, the synthesized images inevitably appear some artifacts. For example, there are some artifacts in the outlines of the two profile faces in the bottom left.
Fig.~\ref{fig-celeba} (b) presents the results of multiple manipulations of one input face. The above visualization results demonstrate the superior performance of our method in the uncontrolled environment.

\begin{figure*}[t]
    \centering
    \includegraphics[width=0.98\textwidth]{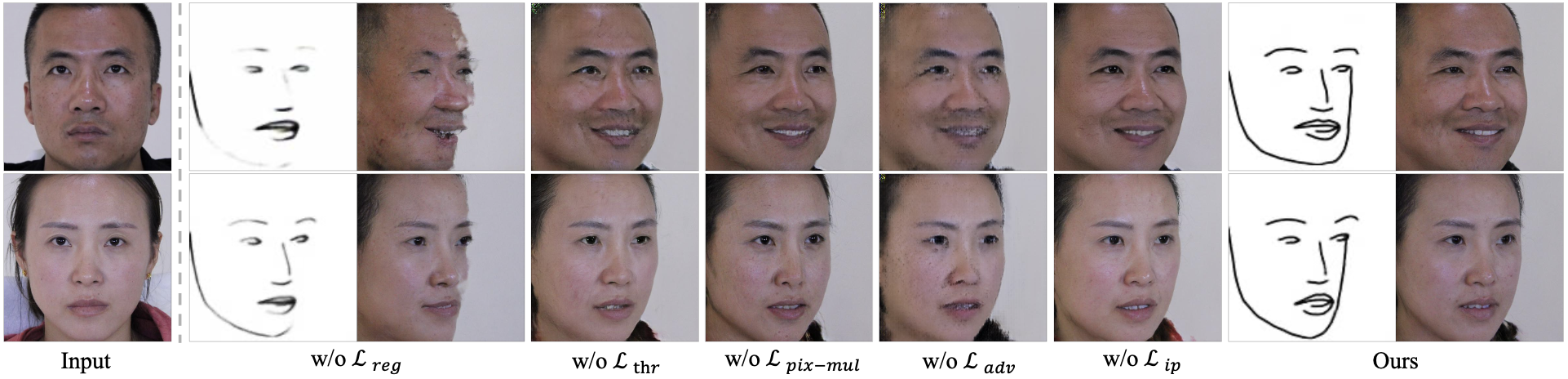}
    \caption{Visualization results (512$\times$512 resolution) of the ablation study: comparisons between our method and its five variants. Please zoom in for details.
    }
    \label{fig-ablation}
\end{figure*}

\subsection{Quantitative Experiments} \label{quantitative-experiments}
In this subsection, we quantitatively evaluate the identity preserving property and the synthesis quality of our method. As shown in Fig.~\ref{fig-multipie}, our method effectively recovers structures and textures from the profiles, which can be used to improve the performance of face recognition under large poses \cite{hu2018pose}. Therefore, we compare the recognition accuracy of our method with that of the state-of-the-art face frontalization methods, including DA-GAN~\cite{yin2020dual}, 3D-PIM~\cite{zhao20183d}, CAPG-GAN~\cite{hu2018pose}, PIM~\cite{zhao2018towards}, TP-GAN~\cite{huang2017beyond}, FF-GAN~\cite{yin2017towards}, and DR-GAN~\cite{tran2017disentangled} on the MultiPIE Setting $2$ protocol. The probe set consists of profiles with various views and the gallery set only contains one frontal face per subject. The profiles in the probe set are frontalized by the above methods, and the pre-trained LightCNN \cite{wu2018light} is used to extract features. Cosine distances are calculated as the similarities to obtain the Rank-1 accuracies, the comparisons of which are tabulated in Table~\ref{table-1}. `LightCNN' means evaluation on the original profiles via the pre-trained LightCNN model. `Ours' means calculating the Rank-1 accuracy on the synthesized frontal faces by the same LightCNN model. We observe that as the face angle increases, the accuracies of all the methods drop gradually. The degradation may be caused by the loss of facial appearance of profiles. Furthermore, the Rank-1 accuracies of all the methods are comparable under small poses ($\pm 15^o$, $\pm 30^o$, and $\pm 45^o$). But at the larger poses, the superiority of our method is obvious. Particularly, our method significantly improves the accuracy under the challenging $\pm 90^o$, and obtains the best performance compared with other state-of-the-art methods.

\begin{table}[t]
    \centering
    \caption{Comparisons of Rank-1 recognition rates (\%) across views under the MultiPIE Setting 2.}
    \label{table-1}
    \resizebox{0.49\textwidth}{!}{
        \begin{tabular}{lcccccc}
        \toprule[0.9pt]
        Method & $\pm 15^o$ & $\pm 30^o$ & $\pm 45^o$ & $\pm 60^o$ & $\pm 75^o$ & $\pm 90^o$ \\
        \midrule[0.8pt]
        DR-GAN~\cite{tran2017disentangled}& $94.0$ & $90.1$ & $86.2$ & $83.2$ & - & - \\
        FF-GAN~\cite{yin2017towards} & $94.6$ & $92.5$ & $89.7$ & $85.2$ & $77.2$ & $61.2$ \\
        TP-GAN~\cite{huang2017beyond} & $98.6$ & $98.0$ & $95.3$ & $87.7$ & $77.4$ & $64.6$ \\
        PIM~\cite{zhao2018towards} & $99.3$ & $99.0$ & $98.5$ & $98.1$ & $95.0$ & $86.5$ \\
        CAPG-GAN~\cite{hu2018pose} & $99.8$ & $99.5$ & $97.3$ & $90.6$ & $83.0$ & $66.0$ \\
        3D-PIM~\cite{zhao20183d} & $99.6$ & $99.4$ & $98.8$ & $98.3$ & $95.2$ & $86.7$ \\
        DA-GAN~\cite{yin2020dual} & \bm{$99.9$} & $99.8$ & $99.1$ & $97.2$ & $93.2$ & $81.5$ \\
        \midrule
        LightCNN~\cite{wu2018light} & $98.5$ & $97.3$ & $92.1$ & $62.0$ & $24.1$ & $5.5$ \\
        Ours & \bm{$99.9$} & \bm{$99.9$} & \bm{$99.4$} & \bm{$98.7$} & \bm{$96.3$} & \bm{$87.4$} \\
        \bottomrule[0.9pt]
        \end{tabular}
    }
\end{table}

\begin{table}[t]
        \centering
        \caption{Comparisons of FID (lower is better) and Rank-1 recognition rates (\%) on the MVF-HQ database.}
        \label{table-2}
        \resizebox{0.49\textwidth}{!}{
            \begin{tabular}{lccccccc}
            \toprule[0.9pt]
            Method & FID & $\pm 15^o$ & $\pm 30^o$ & $\pm 45^o$ & $\pm 60^o$ & $\pm 75^o$ & $\pm 90^o$ \\
            \midrule[0.8pt]
            CAPG-GAN~\cite{hu2018pose} & 36.68 & $99.9$ & $99.9$ & $99.4$ & $95.2$ & $82.1$ & $53.7$ \\
            pix2pixHD~\cite{wang2018high} & 45.62 & $98.5$ & $97.5$ & $93.3$ & $86.1$ & $67.4$ & $39.0$ \\
            pix2pixHD+boundary & 43.37 & $99.2$ & $98.1$ & $94.5$ & $87.3$ & $68.7$ & $40.3$ \\
            \midrule
            LightCNN~\cite{wu2018light} & - & \bm{$100$} & \bm{$100$} & \bm{$99.6$} & $95.7$ & $65.2$ & $23.9$ \\
            Ours & \bm{$12.94$} & \bm{$100$} & \bm{$100$} & \bm{$99.6$} & \bm{$96.5$} & \bm{$84.6$} & \bm{$60.4$} \\
            \bottomrule[0.9pt]
            \end{tabular}
        }
\end{table}

The settings of the probe and the gallery of our MVF-HQ database are analogous to the MultiPIE database. Table~\ref{table-2} shows the comparison results of our method against other state-of-the-art methods, including pix2pixHD (`pix2pixHD+boundary' denotes concatenating the face image with a boundary image, which is generated by the boundary prediction stage, as the input of pix2pixHD) and CAPG-GAN. It is obvious that our method outperforms its competitors by a large margin under the extreme poses ($\pm 75^o$ and $\pm 90^o$). We owe the significant improvements to the introduced proxy network and the feature threshold loss, which disentangle structures and textures in the latent space. The quantitative recognition results are consistent with the qualitative visualization results in Fig.~\ref{fig-multipie-compare} and Fig.~\ref{fig-nd-rafd-compare}. The recognition results on MultiPIE and MVF-HQ prove that our method can effectively improve the recognition performance under large poses. 

Furthermore, in order to evaluate the quality of the synthesized images, we compare the Fr$\acute{\text{e}}$chet Inception Distance (FID) results \cite{heusel2017gans} with CAPG-GAN and pix2pixHD. FID is used to measure the distance between the real faces and the synthesized faces. The results in Table~\ref{table-2} qualitatively reveal the high-quality synthesis character of our method.

\begin{table}[t]
        \centering
        \caption{Quantitative results of the ablation study on the MVF-HQ database.}
        \label{table-3}
        \resizebox{0.49\textwidth}{!}{
            \begin{tabular}{lccccccc}
            \toprule[0.9pt]
            Method & FID & $\pm 15^o$ & $\pm 30^o$ & $\pm 45^o$ & $\pm 60^o$ & $\pm 75^o$ & $\pm 90^o$ \\
            \midrule[0.8pt]
            w/o $\mathcal{L}_{\text{thr}}$ & 36.71      & $96.2$       & $90.7$      & $77.3$      & $67.1$         & $55.6$ & $47.6$ \\
            w/o $\mathcal{L}_{\text{adv}}$ & 53.92      & $98.5$       & $96.1$      & $87.9$      & $80.3$         & $68.4$ & $50.3$ \\
            w/o $\mathcal{L}_{\text{ip}}$ & 21.34       & $88.7$       & $81.5$      & $75.3$      & $65.4$         & $50.7$ & $45.5$ \\
            w/o $\mathcal{L}_{\text{pix-mul}}$ & 34.34  & $98.3$   & $94.3$  & $82.4$  & $81.6$     & $71.3$ & $55.7$ \\
            \midrule
            Ours & \bm{$12.94$} & \bm{$100$} & \bm{$100$}                 & \bm{$99.6$}            & \bm{$96.5$} & \bm{$84.6$}     & \bm{$60.4$} \\
            \bottomrule[0.9pt]
            \end{tabular}
        }
\end{table}

\begin{figure*}[t]
    \centering
    \includegraphics[width=0.98\textwidth]{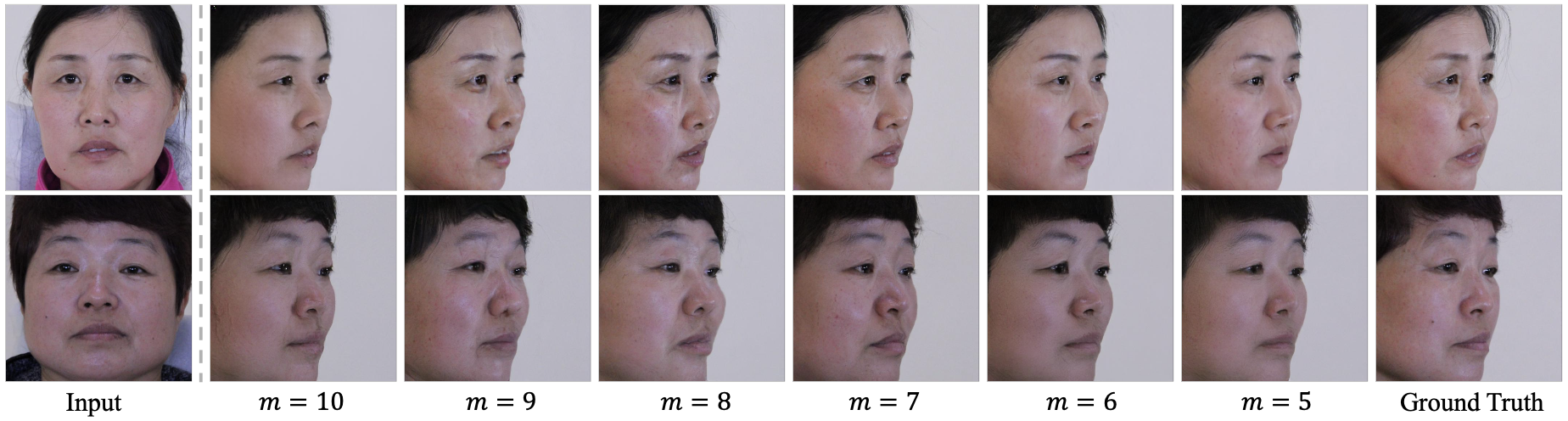}
    \caption{Visualization results (512$\times$512 resolution) of the parameter analysis of $m$ in Eq.~(\ref{eq:threshold}). Please zoom in for details.
    }
    \label{fig-parameter}
\end{figure*}

\subsection{Experimental Analysis} \label{experimental-analysis}
\textbf{Ablation Study.}
In this part, we investigate the roles of the five loss functions in our method, including the conditional regression loss $\mathcal{L}_{\text{reg}}$ in Eq.~(\ref{eq:regression}), the feature threshold loss $\mathcal{L}_{\text{thr}}$ in Eq.~(\ref{eq:threshold}), the multi-scale pixel-wise $\mathcal{L}_{\text{pix-mul}}$ loss in Eq.~(\ref{eq:pix-2}), the conditional adversarial loss $\mathcal{L}_{\text{adv}}$ in Eq.~(\ref{eq:adv}), and the identity preserving loss $\mathcal{L}_{\text{ip}}$ in Eq.~(\ref{eq:ip}).
Both qualitative and quantitative experimental results are reported for better comparisons.

Fig.~\ref{fig-ablation} shows the qualitative visualization results of our method and its five variants.
We discover that without $\mathcal{L}_{\text{reg}}$, the generated boundary images are unsatisfactory, when the given conditional vectors are not completely consistent with those in the databases.
The outlines of many facial components, such as the nose and jaw, are unclear, resulting in incomplete synthesized faces. 
It demonstrates the effectiveness of the conditional regression loss $\mathcal{L}_{\text{reg}}$.
Without the feature threshold loss $\mathcal{L}_{\text{thr}}$, the local structures, \textit{e.g.} the eyes and nose, are unclear and the textures are blurry, indicating the effectiveness of the disentanglement. Moreover, the parameter $m$ in $\mathcal{L}_{\text{thr}}$ has non-negligible impacts on the synthesis results, which will be discussed in the following part. Without the multi-scale pixel-wise loss $\mathcal{L}_{\text{pix-mul}}$ (only utilizing one scale pixel-wise loss), the global structure is clear but the local textures, \textit{e.g.} the teeth, are blurry. Hence, the multi-scale pixel-wise loss $\mathcal{L}_{\text{pix-mul}}$ contributes to the recovery of texture details. Without $\mathcal{L}_{\text{adv}}$, there are many artifacts in the synthesized images, revealing the validity of the multi-scale conditional adversarial loss. Without $\mathcal{L}_{\text{ip}}$, the local textures, such as the beard, are somewhat light. Thus, the identity preserving loss may be beneficial to the enhancement of local textures.

Table~\ref{table-3} further tabulates the FID and Rank-1 results of different variants of our method.
We observe that the FID will increase and the Rank-1 will decrease if one loss is not adopted, which are consistent with the qualitative visualization results in Fig.~\ref{fig-ablation}. These qualitative and quantitative results verify that each component in our method is essential for the high-resolution extreme face manipulation.

\begin{figure}[t]
    \centering
    \includegraphics[width=0.485\textwidth]{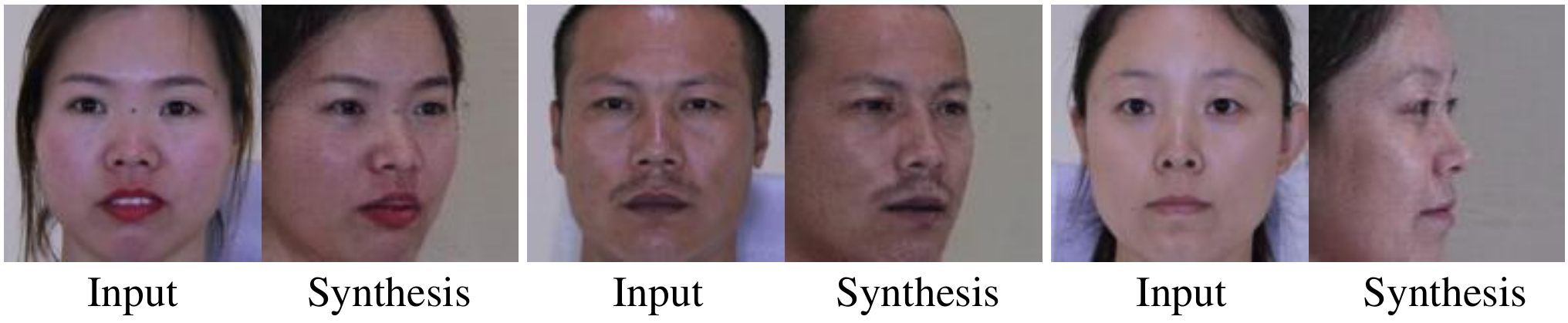}
    \caption{Synthesis results of the cross-database experiment (trained on MultiPIE and tested on MVF-HQ).
    }
    \label{fig-cross}
\end{figure}

\begin{table}[t]
    \centering
    \caption{FID and Rank-1 accuracies ($\pm 90^o$) on the MVF-HQ database under different values of $m$ in Eq.~(\ref{eq:threshold}).}
    \label{table-5}
    \resizebox{0.48\textwidth}{!}{
        \begin{tabular}{lcccccc}
        \toprule[0.8pt]
        $m$ & 10 & 9 & 8 & 7 & 6 & 5 \\
        \midrule[0.6pt]
        FID & $38.65$ & $23.62$ & $17.63$ & \bm{$12.94$} & $20.35$ & $35.43$ \\
        Rank-1 & $48.2$ & $54.7$ & $57.3$ & \bm{$60.4$} & $56.6$ & $50.7$ \\
        \bottomrule[0.8pt]
        \end{tabular}}
\end{table}

\textbf{Cross-Database Experiments.}
Fig.~\ref{fig-cross} plots the results of cross-database experiments. That is, the model is trained on the MultiPIE database and tested on the MVF-HQ database. There is a large domain gap between the two databases, because of the differences in the acquisition equipment, participants, backgrounds, etc. Although the synthesized images on the MVF-HQ database inevitably bring some domain information of the MultiPIE database, such as the backgrounds, our method successfully manipulates the input faces. The results of cross-database experiments further demonstrate the generalization ability of our method.

\textbf{Parameter Analysis.} \label{parameter-analysis}
As mentioned in Section \ref{feature-threshold-Loss}, the value of the parameter $m$ in the feature threshold loss $\mathcal{L}_{\text{thr}}$ (Eq.~(\ref{eq:threshold})) has non-negligible effects on the disentanglement. In Fig.~\ref{fig-parameter}, we plot the visualization results under different values. We observe that when the value of $m$ is too large, the synthesized faces are blurry due to the weak disentanglement. On the contrary, when the value of $m$ is too small, the textures of the synthesized faces will be somewhat lost because of the too compact texture features. The best results are obtained when $m=7$. In addition, the quantitative FID and Rank-1 results are listed in Table~\ref{table-5}. The results of these quantitative indicators are consistent with the visualization results in Fig.~\ref{fig-parameter}. When $m$ equals $7$, we obtain the minimum FID and the highest accuracy.

\section{Conclusion}
This paper has developed a stage-wise framework for high-resolution extreme face manipulation. It simplifies the face manipulation into two correlated stages: a boundary prediction stage and a disentangled face synthesis stage. The first stage predicts the boundary image of the target face in a semi-supervised way. The second stage utilizes the predicted boundary to perform realistic face synthesis. A proxy network and a feature threshold loss are introduced to disentangle structures and textures in the latent space. Furthermore, a new high-resolution MVF-HQ database has been created, which consists of 120,283 images at 6000$\times$4000 resolution from 479 identities. It is much larger in scale and much higher in resolution than publicly available high-resolution face manipulation databases. 
In the future, we will continue to collect more data to further enrich MVF-HQ.
Extensive experiments on four databases show that our method significantly pushes forward the advance of extreme face manipulation.

\section*{Acknowledgments}
The authors would like to greatly thank the associate editor and the reviewers for their valuable comments and advice. This work is partially funded by Beijing Natural Science Foundation (Grant No. JQ18017) and National Natural Science Foundation of China (Grant No. 61721004, U20A20223).

\ifCLASSOPTIONcaptionsoff
  \newpage
\fi

{
\bibliographystyle{IEEEtran}
\bibliography{mybibfile}
}

\end{document}